%% file: acl_latex.tex
\documentclass[11pt]{article}

\usepackage[final]{acl}

\usepackage{times}
\usepackage{latexsym}

\usepackage[T1]{fontenc}

\usepackage[utf8]{inputenc}

\usepackage{microtype}

\usepackage{inconsolata}
\usepackage{amsfonts}

\usepackage{graphicx}

\usepackage{enumitem}
\usepackage{algorithm}
\usepackage{algorithmicx}
\usepackage{algpseudocode}
\usepackage{amsmath} 
\usepackage[table]{xcolor}

\usepackage{booktabs}   
\usepackage{arydshln}   
\usepackage{tcolorbox}
\tcbuselibrary{breakable}
\usepackage{xcolor}
\usepackage{multirow}
\usepackage{makecell}
\usepackage{placeins}

\usepackage{tikz}

\title{FAMA: \underline{F}ailure-\underline{A}ware \underline{M}eta-\underline{A}gentic Framework for Open-Source LLMs in Interactive Tool Use Environments}

\author{First Author \\
  Affiliation / Address line 1 \\
  Affiliation / Address line 2 \\
  Affiliation / Address line 3 \\
  \texttt{email@domain} \\\And
  Second Author \\
  Affiliation / Address line 1 \\
  Affiliation / Address line 2 \\
  Affiliation / Address line 3 \\
  \texttt{email@domain} \\}

\author{Amir Saeidi$^{\dagger*}$ \quad Venkatesh Mishra$^\dagger$\thanks{\ Equal Contribution} \quad Souradeep Mukhopadhyay$^\dagger$  \\  \textbf{Gaowen Liu}$^\ddagger$ \quad \textbf{Ali Payani}$^\ddagger$ \quad  \textbf{Jayanth Srinivasa}$^\ddagger$ \quad \textbf{Chitta Baral}$^\dagger$ \\\\ 
$^\dagger$Arizona State University \quad $^\ddagger$Cisco Research\\
\small{\texttt{\{ssaeidi1, vmishr23, chitta\}@asu.edu,}}
\small{\texttt{\{gaoliu, apayani, jasriniv\}@cisco.com}}
}

\begin{document}
\maketitle

\input{sections/1_abstract}
\input{sections/2_introduction}
\input{sections/3_related_work}
\input{sections/4_problem_statement}
\input{sections/5_method}
\input{sections/6_experiments}

\input{sections/8_conclusion}



\section*{Limitations}

While the proposed Failure-Aware Meta-Agentic (FAMA) framework achieves consistent improvements across multiple benchmarks, several limitations remain.

First, FAMA operates over a predefined pool of specialized agents inherited from the underlying agentic framework (§4.1). This design reflects a common setting in existing user-facing, modular tool-use systems, where a fixed set of components is orchestrated rather than dynamically generated. As a result, FAMA’s effectiveness is bounded by the coverage of this agent pool: failure modes that are not well captured by existing agents may not be adequately mitigated. Importantly, this assumption does not imply rigidity. FAMA is designed to be extensible. New specialized agents targeting emerging failure patterns can be incorporated without modifying the orchestration mechanism. However, the current framework does not address the automatic discovery or synthesis of such agents, which remains an important direction for future work.

Furthermore, the benchmarks considered in this study primarily focus on structured conversational environments. While these settings provide a controlled testbed for analyzing failure-aware orchestration, they do not capture the full spectrum of interactive agent deployments. Extending FAMA to more diverse settings, including embodied, multimodal, or open-ended environments with less well-defined failure taxonomies, is an important avenue for future investigation.

\section*{Ethics Statement}

We have utilized AI assistants, specifically Grammarly and ChatGPT, to correct grammatical errors and rephrase sentences.


\section*{Acknowledgement}

We thank the anonymous reviewers for their constructive suggestions. We extend our gratitude to the Research Computing (RC), and Enterprise Technology at ASU for providing computing resources, and access to the GPT API version for experiments. This work was in part supported by a gift award from Cisco Research. This work makes use of the $\tau$-bench, $\tau$-trait, and ACEBench dataset, which are distributed under the MIT License.

\bibliography{custom}

\input{sections/appendix}
\end{document}

%% file: sections/1_abstract.tex
\begin{abstract}
Large Language Models are being increasingly deployed as the decision-making core of autonomous agents capable of effecting change in external environments. Yet, in conversational benchmarks, which simulate real-world customer-centric issue resolution scenarios, these agents frequently fail due to the cascading effects of incorrect decision-making. These challenges are particularly pronounced for open-source LLMs with smaller parameter sizes, limited context windows, and constrained inference budgets, which contribute to increased error accumulation in agentic settings. To tackle these challenges, we present the \textbf{Failure-Aware Meta-Agentic (FAMA)} framework. FAMA operates in two stages: first, it analyzes failure trajectories from baseline agents to identify the most prevalent errors; second, it employs an orchestration mechanism that activates a minimal subset of specialized agents tailored to address these failures by injecting a targeted context for the tool-use agent before the decision-making step. 
Experiments across open-source LLMs demonstrate performance gains up to\textbf{$~$27\%} across evaluation modes over standard baselines. These results highlight that targeted curation of context through specialized agents to address common failures is a valuable design principle for building reliable, multi-turn tool-use LLM agents that simulate real-world conversational scenarios.
\end{abstract}




%% file: sections/2_introduction.tex
\section{Introduction}

Recent advancements in large language models (LLMs) \citep{liu2025deepseek, anthropic2025claudeopus45, deepmind2025gemini3pro, openai2025gpt52} have significantly enhanced their potential to function as the decision-making backbone of autonomous agents \citep{chowa2025language}. Such agents are now being tested across a broad spectrum of applications, ranging from customer support and enterprise automation \citep{KEMELL2025107805,bcg2025agenticaitransforming, zendesk2025buildingmultiturntests} to scientific discovery \citep{zhou2025autonomous, kumbhar-etal-2025-hypothesis}, software engineering \citep{dong2025survey}, and data analysis \citep{rahman2025llm}, where these systems must 
reason over long conversational horizons and generate tokens which form the basis of decisions and actions with real effects on external environments. One of the most important use-cases of these agents is of multi-turn conversational, interactive chatbots, which have augmented external API (also termed as a `tool') invocation capabilities \citep{wang-etal-2025-rethinking-stateful, zhang2025survey, farn2023tooltalk}. Despite impressive progress, recent benchmarks such as $\tau$-bench \citep{yao2024tau}, $\tau$-trait \citep{he2025impatient} and ACEBench \citep{chen2025acebench} reveal that even strong LLM-based agents struggle to reliably and efficiently complete realistic, customer-centric tasks that require sustained reasoning and structured tool use.

\begin{figure}
    \centering
    \includegraphics[width=1\linewidth]{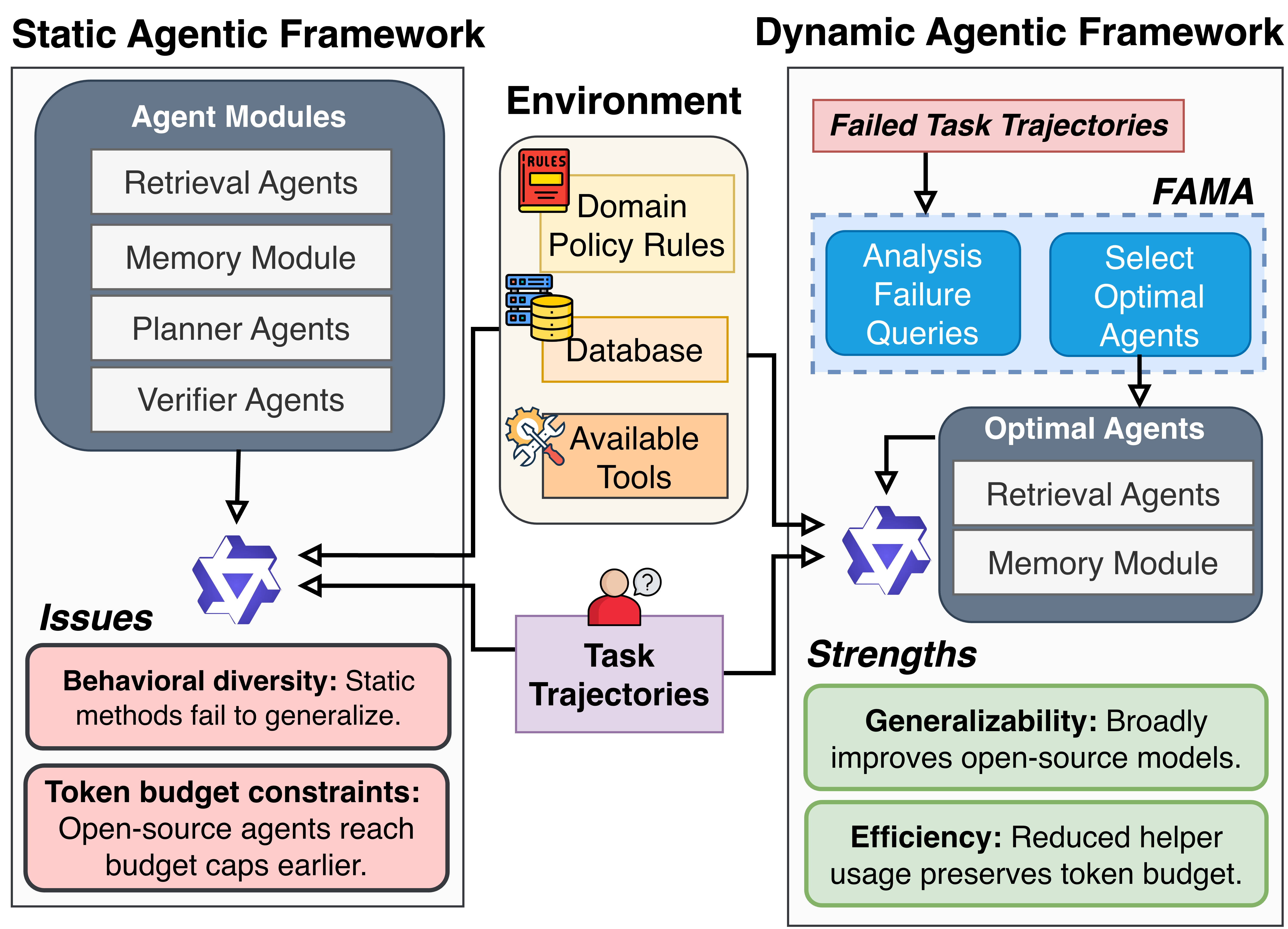}
    \caption{Comparison of static and dynamic (e.g., FAMA) agentic orchestration.}
    \label{fig:fama-teaser}
\end{figure}

\begin{figure*}[t]
    \centering
    \includegraphics[width=1\linewidth]{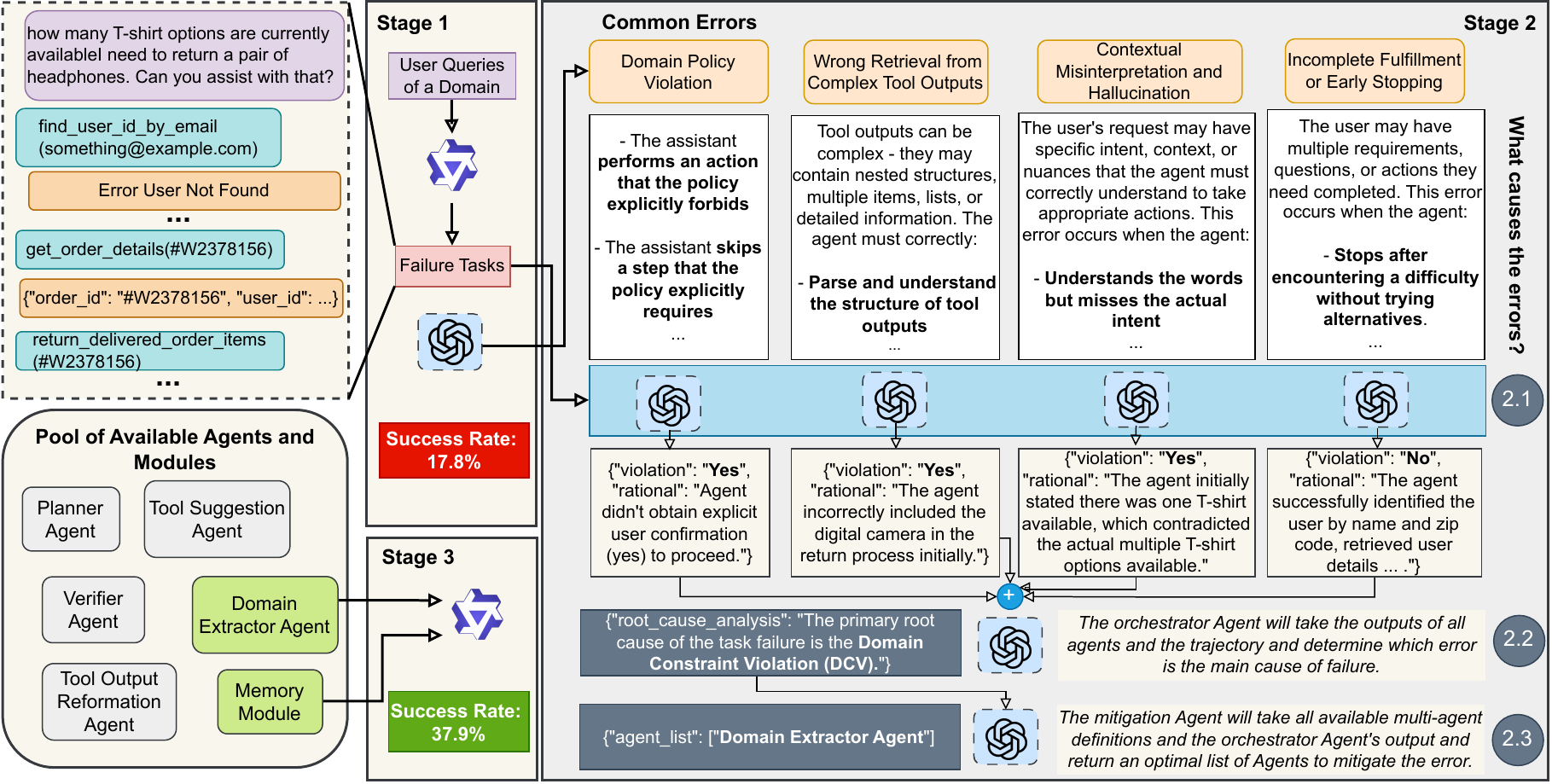}
    \caption{Overview of the FAMA framework. \textbf{Stage 1:} A baseline agent, without multi-agent assistance, is executed on domain tasks. \textbf{Stage 2:} failed tasks are analyzed and categorized by a human or an agentic framework. \textbf{Stage 2.1:} Independent agents focus on distinct issues and analyze the causes of failure through task-specific conversations. \textbf{Stage 2.2:} An orchestrator agent identifies the primary reasons for failure. \textbf{Stage 2.3:} A mitigation agent selects the most appropriate helper agents to address the identified issues. \textbf{Stage 3:} the baseline agent is re-executed using only the helper agents recommended by the mitigation agent.}
    \label{fig:fama_main_figure}
\end{figure*}

A central challenge exposed by tool-use benchmarks is the accumulation of decision-making errors over long trajectories \citep{sinha2025illusion, Mishra2025HowCI, lichkovski2025eu} that involve heterogeneous context, including tool outputs, domain rules, inter-agent interventions, ambiguous user intent and evolving task constraints
\citep{sinha2025illusion, handa2025optagent, lichkovski2025eu, uddin2026recall}. Our experiments (\textsection\ref{subsec:experimental-results}) show that such errors are heterogeneous: different model backbones, particularly open-source models \citep{yang2025qwen3, abdin2024phi} with limited context windows, exhibit distinct dominant failure modes, indicating that static prompting strategies or agentic scaffolding architectures are insufficient.

Prior work has largely sought to improve tool-using agents through task-specific supervised fine-tuning \citep{chen2023fireact, zeng-etal-2024-agenttuning}, or reinforcement learning–based optimization \citep{acikgoz2025speakrl, Zhao2025MUARLMU, jiang2025verltool}. 
For multi-turn tool-use/tool-calling tasks, where trajectories are lengthy, partially observable, and highly variable \citep{kate2025longfunceval, wang2025rethinking}: collecting sufficient high-quality supervision or reward-aligned experience becomes prohibitively expensive \citep{saeidi2026vulcan, golubev2025training, qiwebrl, chen2025scaling}. Reinforcement learning methods \citep{zhang2025landscape} are not overall effective for these challenges, as training requires large-scale curated trajectories, repeated execution of complex tool interactions and long episodes to propagate sparse and delayed rewards, leading to substantial compute and engineering overhead \citep{wang2023mint, wang2025beyond}. Moreover, because training objectives are typically optimized over learning patterns of correct behavior, these methods rarely target the specific failure patterns exhibited by a given agent \citep{NEURIPS2024_cc4d9cfc}, limiting their effectiveness in addressing systematic errors observed in practice.

We have also observed that there has been limited exploration \citep{nekoei2025just, fu2024autoguide} of training-free improvement techniques that extract failure signals from agent trajectories and use them to modularly orchestrate agentic systems built specifically on smaller open-source language models \citep{belcak2025small}. While larger models benefit from extensive pretraining and long context windows, their deployment is often impractical in real-world settings due to cost, latency, and privacy constraints \citep{liang2025llm}. Moreover, scaling alone can mask, rather than resolve, fundamental deficiencies in long-horizon decision-making and tool use. Small language models expose these limitations more clearly \citep{shen-etal-2024-small, sinha2025illusion}, making them an ideal testbed for studying systematic agent failures under realistic resource constraints.

In this work, we argue that reliable multi-turn tool use requires failure-aware agent composition. We introduce a two-stage framework, named the \textbf{Failure-Aware Meta-Agentic (FAMA)} framework, that explicitly identifies the most prevalent failure modes in language agent conversation trajectories and dynamically adapts the tool-use agent’s internal prior context to mitigate those failures. FAMA operates in two stages. First, it analyzes failure trajectories produced by a baseline agent and categorizes errors into a small set of common failure types. \textit{Rather than treating all errors equally, this analysis surfaces the dominant errors that most strongly impact performance.} 

In the second stage, the FAMA framework employs a lightweight routing mechanism consisting of an orchestrator and a mitigation agent (Figure \ref{fig:fama_main_figure}) that selectively activates a set of specialized LLM helper-agents designed to address these specific weaknesses (see \textsection \ref{subsec:FAMA}). 
By dynamically constructing the agent’s prior context using only the necessary agents, FAMA avoids the overhead of excessive context injection while directly addressing the agent’s most critical shortcomings (Figure \ref{fig:fama-teaser}). This design motivates our use of the term `\textit{\textbf{meta-agentic}}', as the proposed system does not act in the environment directly, but operates at a higher level by reasoning over agent behaviors, diagnosing their failures, and dynamically composing a minimal set of specialized agents to shape the decision-relevant context of the underlying tool-use agent.

We evaluate FAMA across $\tau$-bench,  $\tau$-trait and ACEBench using a set of open-source LM backbones with parametric sizes ranging from 4-72 billion. Our results demonstrate consistent improvements in tool-use accuracy, trajectory reliability, and context efficiency. FAMA yields noticeable task-success gains of up to \textbf{25}\% in $\tau$-bench, \textbf{27}\% in ACEBench and \textbf{24}\% in $\tau$-trait over baselines for open-source model backbones, highlighting the effectiveness of targeted error mitigation over brute-force context injection.
In summary, our main contributions are:
\begin{enumerate}[noitemsep]
    \item We propose the \textbf{Failure-Aware Meta-Agentic (FAMA)} framework, which operates in two stages: first, identifying the predominant errors of baseline agents, and then dynamically constructing the prior context for the tool-use agent by selectively enabling the appropriate agents designed to mitigate these errors.
    \item We demonstrate that the FAMA framework’s optimized context construction significantly improves tool-use performance of diverse open-source LLM agents across the $\tau$-bench, $\tau$-trait and ACEBench benchmarks.
    \item We provide a comprehensive analysis of the improvements observed from FAMA over other baseline agents. 
\end{enumerate}



%% file: sections/3_related_work.tex
\section{Related Works}
\label{sec:Related_Works}

\paragraph{LLM-based Tool-Use Benchmarks}
While early benchmarks primarily evaluated isolated tool calls in response to single-turn queries \citep{patil2024gorilla, li-etal-2023-api, qin2023toolllm}, recent work has increasingly shifted toward multi-turn, interactive tool-use benchmarks \citep{yao2024tau, Barres20252BenchEC, chen2025acebench, he2025impatient, lu2025toolsandbox, liu2024convbench}. These benchmarks expose long-horizon reasoning challenges for language agents \citep{cemri2025multi,sinha2025illusion} under partial observability, and sustained environment and task constraints, highlighting the need for agent designs that explicitly address failure accumulation over extended interaction trajectories as a valuable direction.

\paragraph{Training-Free Techniques for Tool-Using Agents}

A range of techniques, including structured prompting \cite{yao2023react, wei2022chain}, and modular agent orchestration \citep{Mishra2025HowCI, zhang2025agentorchestra, zhao2025llm}, have been developed to improve language agents without updating model parameters. However, these approaches are predominantly evaluated on strong or proprietary LLM backbones, with large context windows and robust reasoning capabilities \citep{qu2025tool, chowa2025language}. This remains relatively underexplored for smaller open-source models remains comparatively, particularly in tool-use settings \citep{belcak2025small} where limited context, weaker planning, and error accumulation pose significant challenges \citep{shen-etal-2024-small}, and training techniques are dominantly explored \citep{jhandi2025small, zeng-etal-2025-tool, zhiyuan2025advancing}. FAMA attempts to tackle these challenges by specifically focusing on agentic design through failure modes of smaller open-source LLM agents.

\paragraph{Failure-Aware Agentic Orchestration}

Recent work has begun to systematically analyze agent failures in multi-agent LLM systems \citep{cemri2025multi, lu2025exploring}, uncovering diverse failure causes \citep{Mishra2025HowCI, sinha2025illusion} and proposing taxonomies and annotation datasets to enable more principled debugging and evaluation \citep{zhu2025llm, banerjee2025did}. Other efforts automate failure tracing and attribution \citep{zhang2025agentracer, ge2025introducing} to localize error sources.
Although reflective \citep{shinn2023reflexion, madaan2023self} and experiential \citep{zhang2025agent, huang2025environment} feedback techniques exist, and recent episodic guidance/hinting works \citep{nekoei2025just,fu2024autoguide, wangagent} come close in principle, they do not leverage observed failure patterns to dynamically compose model-specific modular agent scaffolds under resource constraints, an angle we explore with our FAMA framework.

%% file: sections/4_problem_statement.tex
\section{Problem Statement}
\label{sec:Problem_Statement}

For simulating ideal testbeds of tool-use in multi-turn conversations, we utilize tasks from benchmarks such as $\tau$-bench \citep{yao2024tau}, $\tau$-trait \citep{he2025impatient}, and ACEBench \citep{chen2025acebench}. These benchmarks are specifically designed to assess the decision-making capabilities of language agents in realistic, multi-turn interaction settings.
In these setups, both the user and the customer-service assistant are simulated by LLMs, enabling a controlled yet dynamic environment for analyzing interactive behavior. The customer-service agent is the language agent that generates the tokens signifying which tools are to be invoked, while following the specific domain policies. As explained in \citet{shekkizhar2025echoing}, the benchmarks used here do not evaluate standard multi-agent settings \cite{stone2000multiagent} as the distinct agents simulating different roles, such as the user or the assistant, can have private internal states and competing utilities. Further details on these benchmarks are provided in Appendix \ref{sec:appendix_benchmark_details}.


%% file: sections/5_method.tex
\section{Method}
Although frontier models in agentic frameworks demonstrate strong performance on complex, dynamic tasks, their deployment is often prohibitively expensive. Additionally, integrating these models into private or proprietary products is often impractical due to cost and security constraints. Open-source models, by contrast, offer a viable alternative; their smaller size enables local deployment, making them more cost-effective and better suited for privacy-sensitive applications.

Recent works \citep{liu2025acemath, rrv-etal-2025-thinktuning, handa2025guidedsampling} show that open-source models, along with inference-time frameworks built around them, can achieve performance comparable to frontier models on mathematical and reasoning tasks; however, their use within agentic frameworks to solve a complex task remains challenging. Due to their smaller size and the nature of the data on which they are fine-tuned, these models exhibit distinct failure modes and limitations. Additionally, strict context constraints limit their applicability in complex multi-agent frameworks, where multiple agents must collaborate to solve a task, and rapidly fill up the context windows. These factors lead open-source models to exhibit weaker decision-making capabilities compared to frontier models.

To address these challenges, we propose the \textbf{Failure-Aware Meta-Agentic (FAMA)} framework, a general and dynamic framework designed to identify failure patterns in the behavior of open-source model agents and determine an effective set of agents that achieves improved performance under resource constraints. In the following sections, we present and analyze the FAMA framework in detail.

\subsection{Failure-Aware Meta-Agentic Framework}
\label{subsec:FAMA}
We assume the availability of a predefined set of agents $\mathcal{A}$, where each individual agent is denoted by $a \in \mathcal{A}$, and subsets of agents are denoted by $\mathcal{A}' \subseteq \mathcal{A}$. The full agent set $\mathcal{A}$ is adapted from \citet{Mishra2025HowCI}, which proposes a collection of specialized agents for improving performance in tool-calling environments, and is further extended with additional modules tailored to the benchmarks evaluated in this work. Specifically, $\mathcal{A}$ includes a Domain Constraints Extractor Agent (DCE), a Tool Suggestion Agent (TSA), a Tool Output Reformulator Agent (TOR), a Planner Agent, a Decision Verifier Agent, and a User Context Manager (Memory) module.

We further assume the availability of a predefined set of general error categories representing common agent failures, denoted by $\mathcal{E}$, where a realized error instance is denoted by $e \in \mathcal{E}$. Based on prior studies \cite{Mishra2025HowCI, shekkizhar2025echoing, kokane2024spectool} and our own analysis, we categorize failures in tool-calling environments into four classes: \textbf{(1)} Domain Policy Violation, \textbf{(2)} Incorrect Retrieval from Complex Tool Outputs, \textbf{(3)} Contextual Misinterpretation and Hallucination, and \textbf{(4)} Incomplete Fulfillment or Early Stopping. Detailed definitions of these error categories are provided in Appendix \ref{appendix:err_cat_def}. 

Given the agent set $\mathcal{A}$ and the error category set $\mathcal{E}$, we define $|\mathcal{E}|$ independent error analysis agents, each responsible for detecting the cause of failure associated with a specific error category $e$. These agents operate over a predefined list of failure causes derived from empirical analysis of baseline trajectories. Additional details on the causes of failure for each error category are provided in Appendix \ref{appendix:err_cat_def}. Each error analysis agent produces both a categorical decision and a supporting rationale in textual form.

The textual outputs and rationales produced by all error analysis agents are concatenated into a single input and, together with the full interaction trajectory between the user and the tool-calling agent for a given task $\tau$, are passed to the orchestrator agent for final failure attribution. This process explicitly accounts for cases in which an agent may initially make incorrect decisions but later recover after receiving feedback from the environment, ensuring that failure attribution reflects the overall trajectory rather than isolated errors.

In the final stage, conditioned on the error categories identified by the orchestrator agent and the functional definitions of agents in $\mathcal{A}$, a dedicated mitigation agent determines an efficient subset of agents $\mathcal{A}^* \subseteq \mathcal{A}$ that can mitigate a specific error category $e$. Aggregating the mitigation agent’s recommendations across the selected error categories yields a final agent subset that represents the minimal configuration required to address the dominant failure modes.

We collect these recommendations across tasks and deploy multi-agent frameworks composed only of the selected agent subsets (See Algorithm \ref{alg:fama}). The results reported in Tables \ref{tab:tau_bench_results}-\ref{tab:tau_trait_results} demonstrate that FAMA consistently achieves better performance across different settings compared to IRMA, ReAct, and FC on both $\tau$-bench, $\tau$-trait, and ACEBench across different models.

\input{tables/fama_algo}

%% file: tables/fama_algo.tex
\begin{algorithm}[t]
\caption{Failure-Aware Meta-Agent (FAMA)}
\label{alg:fama}
\begin{algorithmic}[1]
\Require Agent set $\mathcal{A}$; error categories $\mathcal{E}$; task set $\mathcal{T}=\{\tau_1,\dots,\tau_m\}$
\Ensure Selected agent subsets $\{\mathcal{A}^{\star}_{\tau}\}_{\tau\in\mathcal{T}}$, where $\mathcal{A}^{\star}_{\tau}\subseteq\mathcal{A}$

\State Initialize failure set $\mathcal{F}\gets \emptyset$
\For{each task $\tau \in \mathcal{T}$}
    \State $(\xi_\tau, r_\tau) \gets \textsc{Execute}(\tau)$ \Comment{$\xi_\tau$: full interaction trajectory}
    \If{$r_\tau = 0$}
        \State $\mathcal{F}\gets \mathcal{F}\cup \{(\tau,\xi_\tau)\}$ \Comment{failure case}
    \EndIf
\EndFor

\For{each $(\tau,\xi)\in \mathcal{F}$}
    \For{each error category $e\in\mathcal{E}$}
        \State $o_{\tau,e} \gets \textsc{Analyze}_{e}(\xi)$ \Comment{independent error-analysis agent for $e$}
    \EndFor
    \State $O_{\tau} \gets \textsc{Concat}(\{o_{\tau,e}\}_{e\in\mathcal{E}})$
    \State $\hat{\mathcal{E}}_{\tau} \gets \textsc{Orchestrate}(O_{\tau}, \xi)$ \Comment{main failure error(s)}
    \State $\mathcal{A}^{\star}_{\tau} \gets \textsc{Mitigate}(\hat{\mathcal{E}}_{\tau}, \mathcal{A})$ \Comment{optimal agent subset}
\EndFor

\State \Return $\{\mathcal{A}^{\star}_{\tau}\}_{\tau\in\mathcal{T}}$
\end{algorithmic}
\end{algorithm}

%% file: sections/6_experiments.tex
\section{Experiments}
\subsection{Experimental Setup}
In this section, we mention the models and benchmarks used in our experiments. Our primary focus is on open-source models; accordingly, we evaluate their performance exclusively on tool-calling benchmarks.

\paragraph{Models.}
We evaluate models from the Qwen family, including \textbf{Qwen3-4B-Instruct}, \textbf{Qwen3-14B}, \textbf{Qwen3-32B}, and \textbf{Qwen2.5-72B-Instruct}. We exclude reasoning- or thinking-augmented variants from our study, as they fall outside the scope of this work. Our analysis indicates that such models consume a substantial portion of their token budget in the internal reasoning process, which introduces token limit constraints when deployed within agentic frameworks. As a result, these models tend to exhibit inferior performance in tool-calling settings.

\paragraph{Methods.}
To assess the effectiveness of FAMA, we compare it against several established baselines, including \textbf{Function Calling (FC)}, \textbf{ReAct}, and the \textbf{Input-Reformulation Multi-Agent framework (IRMA)}. The IRMA study also investigates an enhanced variant incorporating a Self-Reflection mechanism, and reports improved performance over the standard IRMA setup. Since this extension is orthogonal to our focus, we do not include it in our experiments and leave its integration with FAMA for future work.

\input{tables/main_res_tau_bench}

\paragraph{Evaluation.} 
In this work, we focus on dynamic environments in which a user-agent simulator interacts with a tool-calling agent. Accordingly, we evaluate all methods on three benchmarks. First, we use the $\tau$-bench, with evaluations conducted in the Airline and Retail domains. Second, we consider $\tau$-trait, where performance is assessed in the Telehealth and Telecom domains. Finally, we evaluate the methods on the single-agent setting of ACEBench, which covers tasks related to food delivery services and telecommunication operations. For detailed information about the benchmarks and evaluation metrics, refer to Appendix \ref{sec:appendix_benchmark_details}.

We divide our experiments into two stages. First, we identify the best-performing open-source model to serve as the user agent. Next, we evaluate different methods across multiple open-source tool-calling models while fixing the user agent to the selected best-performing model. We additionally note that \textbf{GPT-4o} \citep{achiam2023gpt} and \textbf{GPT-4.1-mini} are used as the judgment agents within the FAMA framework to demonstrate the consistency of the judgments across different models (see Appendix \ref{appendix:more_ablation_studies}).

\begin{figure}
    \centering
    \includegraphics[width=1\linewidth]{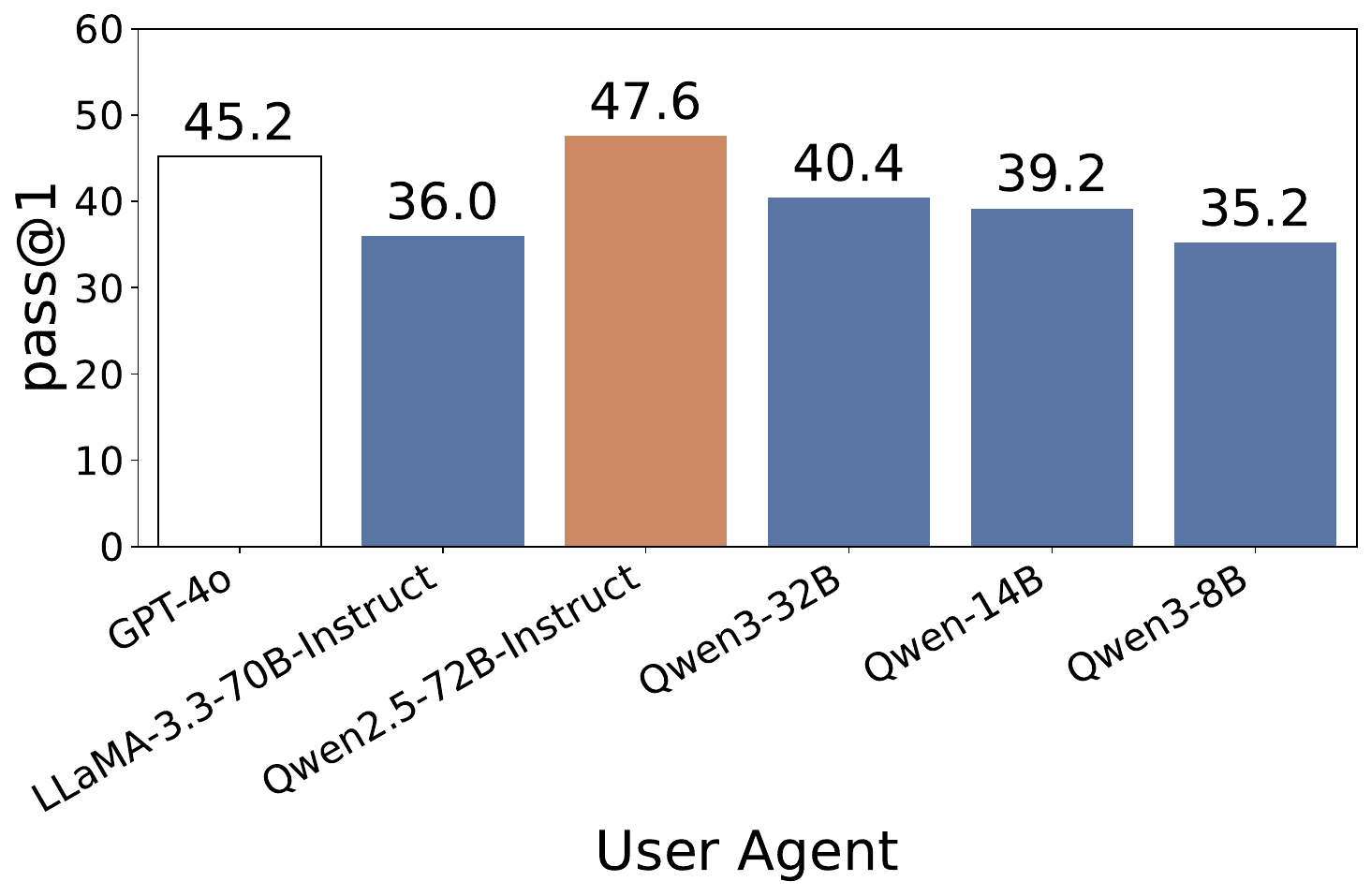}
    \caption{Comparing $\tau$-bench performance across multiple LLMs being tested as user agents.}
    \label{fig:model_pass1_comparison}
\end{figure}

\begin{figure*}[t]
    \centering
    \includegraphics[width=1\linewidth]{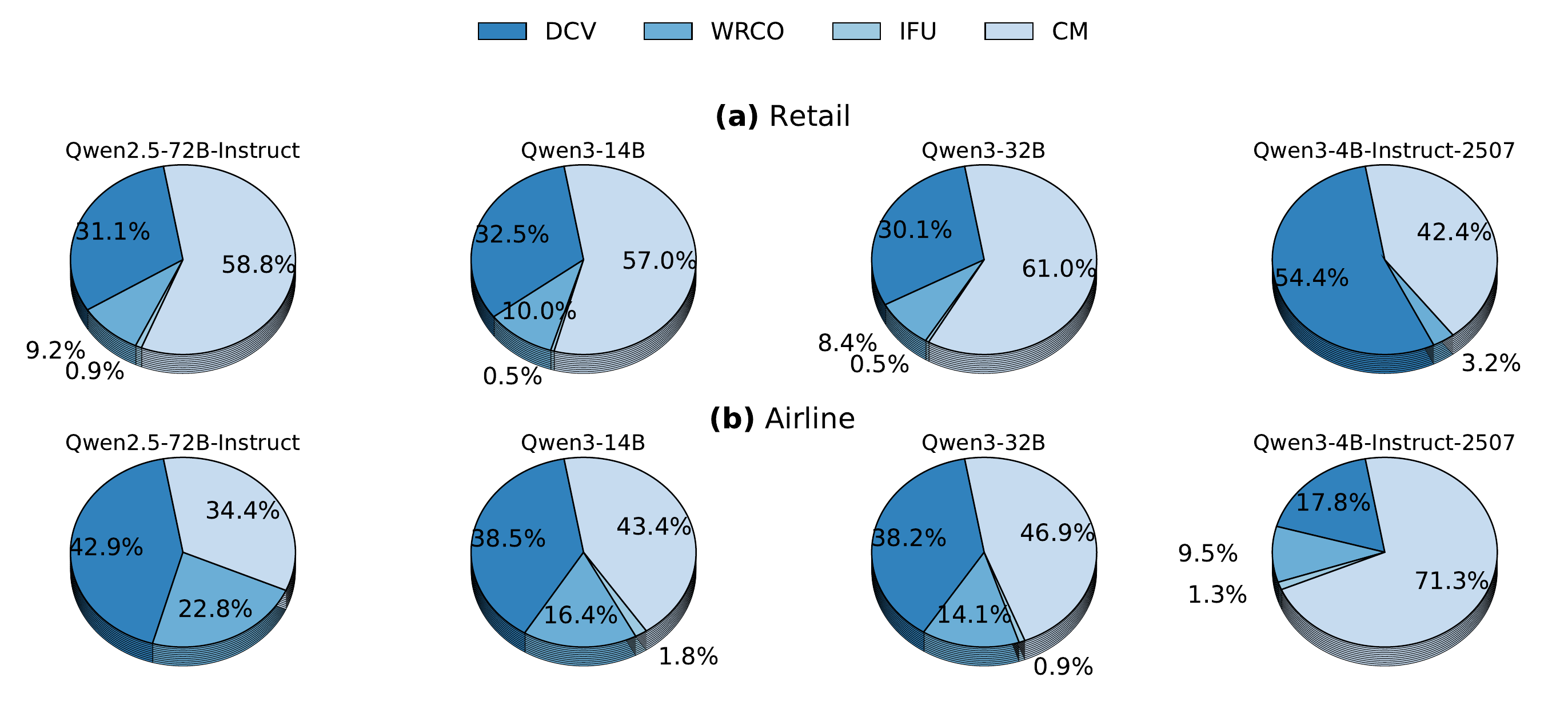}
    \caption{Statistical comparison of error categories across open-source LLMs in $\tau$-retail, as identified by the Orchestrator Agent.}
    \label{fig:error_distribution_tau_bench}
\end{figure*}

\subsection{Experimental Results}
\label{subsec:experimental-results}
\paragraph{Open-source agents can be trusted as user simulator agents.} In this experiment, we use the Qwen3-4B-Thinking model as the tool-calling agent. We select this model due to its relatively small size and its ability to achieve performance comparable to larger models. For the user simulator, we evaluate five state-of-the-art open-source models, namely LLaMA-3.1-70B-Instruct~\citep{grattafiori2024llama}, Qwen2.5-72B-Instruct, Qwen3-8B, Qwen3-14B, and Qwen3-32B \citep{yang2025qwen3}. All experiments are conducted on the Airline domain of $\tau$-bench, and the results are compared against a setting in which GPT-4o is used as the user agent. The results in Figure \ref{fig:model_pass1_comparison} indicate that, among open-source models, Qwen-2.5-72B-Instruct is the closest to a frontier model such as GPT-4o, as its pass@1 performance averaged over five runs surpasses that of GPT-4o.


\paragraph{Using all agents leads to inconsistent performance across models.} Consistent with prior work, we evaluate open-source models on $\tau$-bench in the IRMA setting. Table \ref{tab:tau_bench_results} shows that IRMA outperforms other baselines in a limited number of cases, indicating that while multi-agent frameworks can be beneficial, indiscriminately using all agents does not consistently improve performance across models and can, in some cases, degrade the base agent’s performance.

\paragraph{FAMA consistently outperforms baseline methods across different benchmarks.} Tables \ref{tab:tau_bench_results}–\ref{tab:tau_trait_results} show that the effectiveness of an agentic framework is highly sensitive to both agent selection and domain, and that naively including all agents can degrade performance. By selectively enabling agents based on dominant failure modes, FAMA avoids unnecessary context injection and focuses mitigation on the most impactful errors. As shown in Table \ref{tab:tau_bench_results}, FAMA outperforms ReAct, FC, and IRMA by 4.63\%, 11.57\%, and 5.27\% in the Airline domain and by 5.30\%, 8.96\%, and 6.15\% in the Retail domain, averaged over five runs.


\paragraph{FAMA is a generalizable method.} 
Generalizability is a key limitation of static multi-agent frameworks, which are typically tailored to frontier-model failures in specific domains and require extensive manual redesign to adapt to changes. In contrast, FAMA is a dynamic framework that overcomes this limitation. Experiments across multiple domains and benchmarks, including ACEBench and $\tau$-trait, show that FAMA consistently outperforms other methods by adapting to agent failure behavior (See Tables \ref{tab:ace_bench_results} and \ref{tab:tau_trait_results}). For detailed results on the $\tau$-trait and ACEBench benchmarks, refer to Appendix \ref{appendix:ace_bench_results} and \ref{appendix:tau_trait_results}. 
The modular components in FAMA enable the easy addition of specialized agents to address specific error patterns.

\begin{figure*}
    \centering
    \includegraphics[width=1\linewidth]{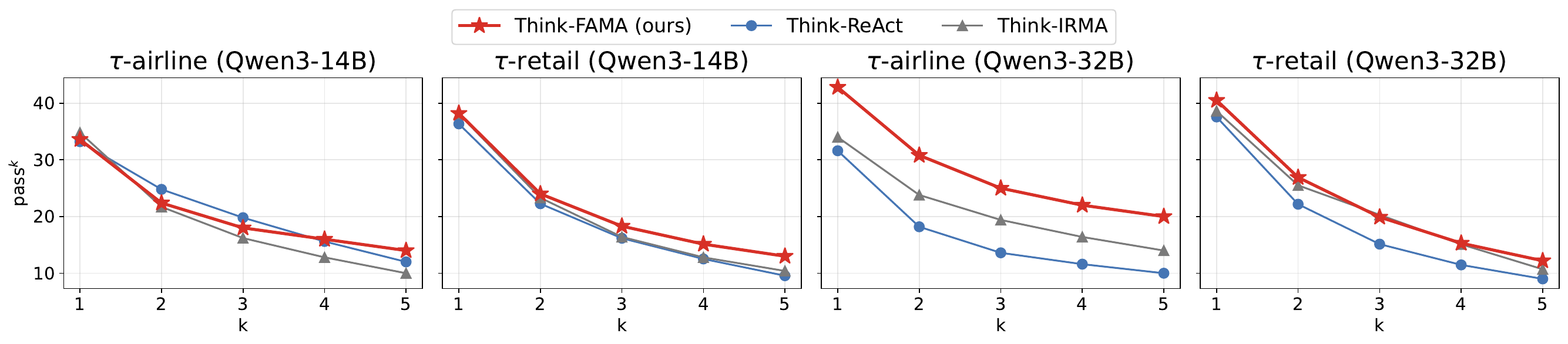}
    \caption{Performance comparison of Qwen3-14B and Qwen3-32B (thinking variants) across methods on $\tau$-bench.}
\label{fig:thinking_variant_qwen3_comparison}
\end{figure*}

To further contextualize our results, we also evaluate reasoning-enabled (“thinking”) variants of Qwen3 models. As shown in Figure \ref{fig:thinking_variant_qwen3_comparison}, while such variants can improve certain baseline methods, they incur higher token usage (see Table \ref{tab:token_latency_overhead_analysis}) and frequently lead to context window overflows in multi-turn settings, where accumulated reasoning traces consume a significant portion of the available token budget. We quantify this effect by counting instances (refer to Table \ref{tab:thinking_model_error_freq}) that exceed the maximum token limit across runs and treat them as failures, as mitigating them would require truncating critical context (e.g., domain constraints or prior tool outputs), potentially leading to incorrect decisions. In contrast, FAMA achieves stronger and more stable performance across k without relying on expensive internal reasoning steps.

\begin{figure*}[t]
    \centering
    \includegraphics[width=1\linewidth]{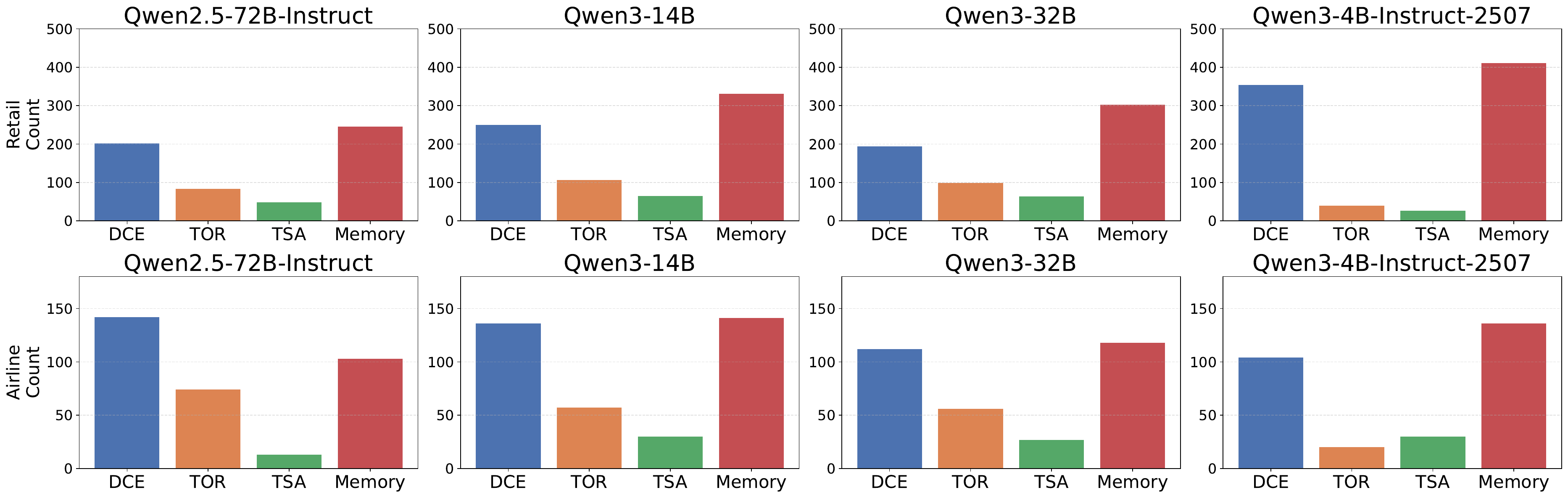}
    \caption{Statistics of recommendations produced by the Mitigation Agent across open source agents for $\tau$-bench.}
    \label{fig:mitigation_agent_tau_bench}
\end{figure*}


\subsection{Analysis and Ablation Study}
To analyze error patterns, we run each task within a domain five times. The results in Figures \ref{fig:error_distribution_tau_bench}, \ref{fig:orch_error_category_tau_trait}, and \ref{fig:orch_error_category_ace_bench} show that all evaluated open-source models exhibit significant difficulty with domain constraint violations and contextual misinterpretations. This indicates that, even when solving a task, selecting actions that strictly adhere to domain-specific rules remains challenging. This issue becomes more pronounced as conversations grow longer, since domain constraints provided in the system prompt tend to be forgotten over time, highlighting memory-related limitations in open-source models. The results in Figure \ref{fig:mitigation_agent_tau_bench} show that the mitigation agent strongly recommends the Memory module and the DCE agent, confirming that memory is a critical bottleneck for open-source agents.


\begin{figure*}[t]
    \centering
    \includegraphics[width=1\linewidth]{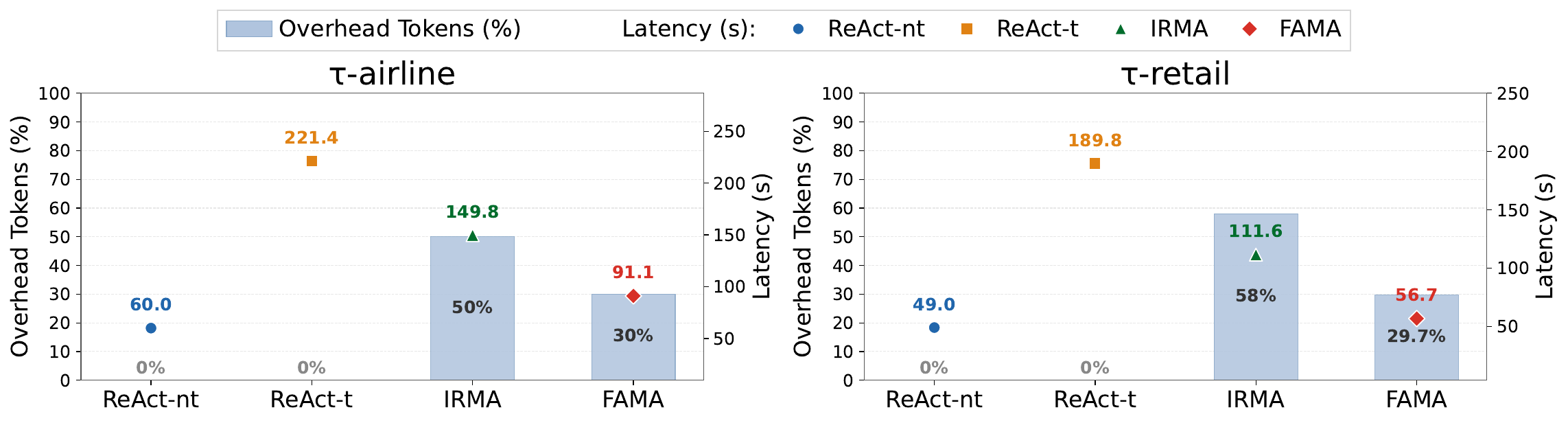}
    \caption{Token overhead (\%) and latency (seconds) across different methods with Qwen3-32B. See Table \ref{tab:token_latency_overhead_analysis} for detailed statistics. `ReAct-nt' and `ReAct-t' represent non-thinking and thinking models used for ReAct method.}
\label{fig:latency_and_overhead_comparison}
\end{figure*}



Motivated by these findings, we conduct an ablation study on memory size to identify effective configurations across domains. As shown in Appendix \ref{sec:optimal_fama_components}, the optimal memory size is domain-dependent rather than model-dependent. Specifically, Figure \ref{fig:ablation_mem_tau_bench} shows that the Retail domain, which involves longer and more complex user–agent interactions, benefits from a larger memory size, with $k=6$ achieving the best performance, whereas the Airline domain attains optimal results with a smaller memory size of $k=2$. We also compare these settings against a no-memory baseline. Overall, the results demonstrate that an agentic framework with an appropriately sized memory not only outperforms IRMA, which uses all agents, but also surpasses other baseline methods. Additional ablation studies on agent combinations that differ from those recommended by the mitigation agent, as well as evaluations with various frontier models in the FAMA framework, have been provided in Appendix \ref{appendix:more_ablation_studies}. A comparison with an additional baseline, Self-Reflection (based on \citet{madaan2023self}), is provided in Appendix \ref{appendix:self_reflection}. 

We also experiment and analyze efficiency in terms of latency and token overhead (see Figure \ref{fig:latency_and_overhead_comparison}) across various baselines and our method. Methods that introduce additional reasoning or helper agents incur higher runtime and token costs, which can worsen context-window size constraints in multi-turn settings. In particular, IRMA exhibits substantial overhead (50 and 58\%), and task completion latency (149.8 and 111.6 seconds averaged) for Qwen3-32B on retail and airline tasks. While FAMA also incurs additional cost compared to ReAct baselines, its more efficient design ($\sim$30\% overhead) results in lower latency than IRMA and fewer overflow-induced errors. Although absolute latency depends on system and deployment configurations, the relative differences across methods are consistent, with IRMA incurring the highest cost. Overall, this suggests that 
the added cost of agentic scaffolding offsets its benefits by straining the available token budget and degrading overall reliability.

%% file: tables/main_res_tau_bench.tex
\begin{table*}[t]
\centering
\small
\begin{tabular}{lrrrrrrrrrr}
\toprule
\textbf{Method} & \multicolumn{5}{c}{\textbf{$\tau$-Airline}} & \multicolumn{5}{c}{\textbf{$\tau$-Retail}} \\
\cmidrule(lr){2-6}\cmidrule(lr){7-11}
& \multicolumn{1}{l}{\textbf{Pass\textasciicircum{}1}}
& \multicolumn{1}{l}{\textbf{Pass\textasciicircum{}2}}
& \multicolumn{1}{l}{\textbf{Pass\textasciicircum{}3}}
& \multicolumn{1}{l}{\textbf{Pass\textasciicircum{}4}}
& \multicolumn{1}{l}{\textbf{Pass\textasciicircum{}5}}
& \multicolumn{1}{l}{\textbf{Pass\textasciicircum{}1}}
& \multicolumn{1}{l}{\textbf{Pass\textasciicircum{}2}}
& \multicolumn{1}{l}{\textbf{Pass\textasciicircum{}3}}
& \multicolumn{1}{l}{\textbf{Pass\textasciicircum{}4}}
& \multicolumn{1}{l}{\textbf{Pass\textasciicircum{}5}} \\
\midrule

\multicolumn{11}{l}{\textbf{\textit{Qwen3-4B-Instruct}}} \\
\addlinespace[3pt]
ReAct & \underline{32.00}\% & \underline{28.00}\% & \underline{26.80}\% & \underline{26.40}\% & \underline{26.00}\%
      & 17.22\% & 12.35\% & 10.61\% & 9.57\% & 8.70\% \\
FC    & 27.60\% & 19.90\% & 17.00\% & 15.20\% & 14.00\%
      & 24.90\% & 16.70\% & 13.00\% & 10.60\% & 9.00\% \\
IRMA  & 30.00\% & 19.60\% & 15.40\% & 13.20\% & 12.00\%
      & \underline{28.86}\% & \underline{17.47}\% & \underline{13.13}\% & \underline{10.95}\% & \underline{9.56}\% \\
FAMA (\textbf{our})  & \textbf{37.60}\% & \textbf{32.00}\% & \textbf{28.30}\% & \textbf{26.70}\% & \textbf{26.00}\%
      & \textbf{34.60}\% & \textbf{24.10}\% & \textbf{19.30}\% & \textbf{16.30}\% & \textbf{13.90}\% \\
\addlinespace[3pt]
\hdashline

\addlinespace[3pt]
\multicolumn{11}{l}{\textit{\textbf{Qwen3-14B}}} \\
\addlinespace[3pt]
ReAct & 18.40\% & 12.00\% & 9.40\% & 8.40\% & 8.00\%
      & 25.20\% & 17.80\% & 14.70\% & 13.20\% & 12.10\% \\
FC    & 14.00\% & 9.60\% & 8.40\% & 8.00\% & 8.00\%
      & \underline{32.60}\% & \underline{22.00}\% & \underline{18.00}\% & \underline{15.00}\% & \underline{13.00}\% \\
IRMA  & \underline{36.40}\% & \textbf{26.40}\% & \textbf{22.00}\% & \textbf{19.60}\% & \textbf{18.00}\%
      & 28.50\% & 15.60\% & 10.80\% & 8.30\% & 6.90\% \\
FAMA (\textbf{our}) & \textbf{36.80}\% & \underline{23.79}\% & \underline{18.20}\% & \underline{16.60}\% & \underline{16.00}\%
      & \textbf{37.90}\% & \textbf{25.70}\% & \textbf{19.70}\% & \textbf{16.30}\% & \textbf{14.70}\% \\
\addlinespace[3pt]
\hdashline

\addlinespace[3pt]
\multicolumn{11}{l}{\textit{\textbf{Qwen3-32B}}} \\
\addlinespace[3pt]
ReAct & \textbf{30.40}\% & \textbf{20.00}\% & \underline{16.20}\% & \underline{14.80}\% & \underline{14.00}\%
      & \underline{38.00}\% & 25.00\% & \underline{18.80}\% & \underline{15.30}\% & 10.00\% \\
FC    & 17.60\% & 13.60\% & 12.00\% & 10.80\% & 10.00\%
      & 35.00\% & \underline{26.60}\% & 15.00\% & 13.10\% & \underline{11.00}\% \\
IRMA  & 24.40\% & \underline{15.90}\% & 12.40\% & 10.00\% & 8.00\%
      & 26.90\% & 15.20\% & 10.90\% & 8.50\% & 6.90\% \\
FAMA (\textbf{our})  & \underline{26.80}\% & \textbf{20.00}\% & \textbf{18.40}\% & \textbf{18.00}\% & \textbf{18.00}\%
      & \textbf{40.50}\% & \textbf{26.90}\% & \textbf{19.90}\% & \textbf{15.30}\% & \textbf{12.20}\% \\
\addlinespace[3pt]
\hdashline

\addlinespace[3pt]
\multicolumn{11}{l}{\textit{\textbf{Qwen2.5-72B-Instruct}}} \\
\addlinespace[3pt]
ReAct & 24.40\% & \underline{18.79}\% & \underline{15.60}\% & \underline{12.80}\% & \underline{10.00}\%
      & \underline{43.47}\% & \underline{32.26}\% & \underline{26.69}\% & \underline{23.13}\% & \underline{20.86}\% \\
FC    & 15.20\% & 7.60\% & 4.00\% & 2.40\% & 2.00\%
      & 19.65\% & 10.34\% & 6.78\% & 5.21\% & 4.34\% \\
IRMA  & \underline{26.40}\% & 16.70\% & 13.60\% & 11.60\% & \underline{10.00}\%
      & 38.78\% & 27.73\% & 22.95\% & 20.34\% & 19.13\% \\
FAMA (\textbf{our})  & \textbf{29.20}\% & \textbf{21.20}\% & \textbf{18.80}\% & \textbf{18.00}\% & \textbf{18.00}\%
      & \textbf{44.17}\% & \textbf{34.26}\% & \textbf{30.26}\% & \textbf{28.17}\% & \textbf{26.95}\% \\

\bottomrule
\end{tabular}
\caption{Comparison of different open-source agents on $\tau$-bench across five runs. Note that Qwen2.5-72B-Instruct is used as the model backbone of the sub-agents in the IRMA and FAMA agentic frameworks.}
\label{tab:tau_bench_results}
\end{table*}

%% file: sections/8_conclusion.tex
\section{Conclusion}
\label{sec:conclusion}

In this work, we present the Failure-Aware Meta-Agentic (FAMA) framework, a dynamic multi-agent approach designed to overcome the limitations of open-source language agents in interactive environments such as tool calling. FAMA operates in two stages: first, it analyzes agent failures to identify dominant error patterns; second, it selects a minimal, task-relevant subset of agents via an orchestrator framework to mitigate these errors. We evaluate FAMA across three interactive tool-calling benchmarks and four open-source LLMs. Results and analyses demonstrate the potential of FAMA as a generalizable, failure-aware orchestration framework for constructing optimized prior contexts as a key design principle for improving and building dependable, resource-efficient conversational agents.


%% file: sections/appendix.tex
\clearpage
\appendix
\section{More Ablation Studies}
\label{appendix:more_ablation_studies}
We conduct additional ablation studies on agent combinations that are not recommended by the mitigation agent, as well as on different judgment models, to assess the quality of the resulting decisions and recommendations.

\input{tables/token_latency_analysis_tau_bench}

\paragraph{Alternative agent combinations exhibit inferior performance.} In this section, we examine the performance of alternative agent combinations that were not recommended by the mitigation agent, in order to further demonstrate the effectiveness of the FAMA method. Specifically, we select Qwen3-14B as the tool-calling agent and conduct Tau-Bench experiments using the following configurations:

\paragraph{Exp 1:} Memory module, Domain Constraints Extractor Agent, and Tool Output Reformulation Agent.
\paragraph{Exp 2:} Memory module and Tool Output Reformulation Agent.
\paragraph{Exp 3:} Memory module, Tool Output Reformulation Agent, and Tool Suggestion Agent.

\begin{figure}[h]
    \centering
    \includegraphics[width=0.92\linewidth]{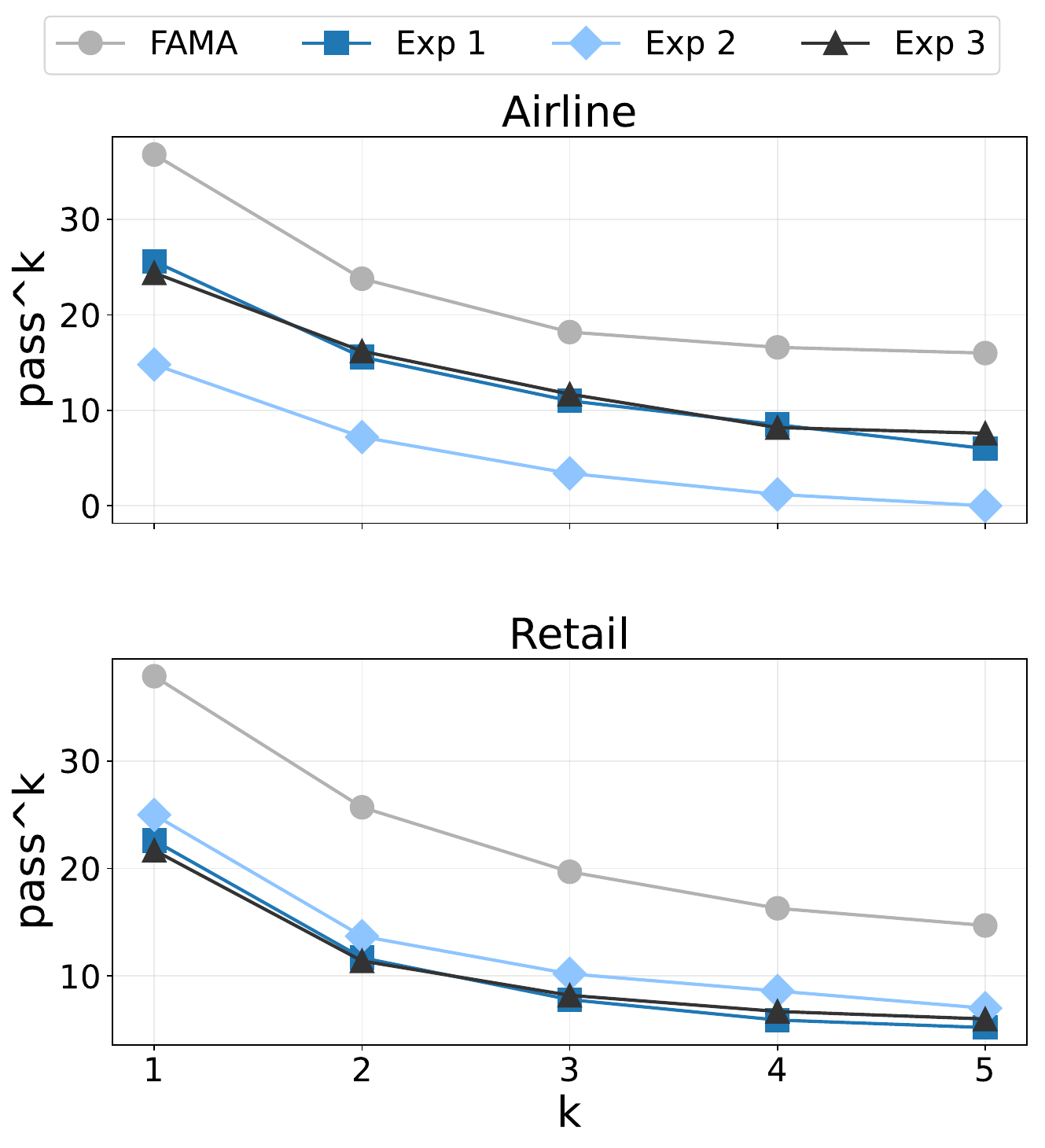}
    \caption{Comparison of tool calling agents that rely on different agentic combinations.}
    \label{fig:ab_diff_com}
\end{figure}

The results in Figure \ref{fig:ab_diff_com} show that the best performance is achieved by the agent combination recommended by the mitigation agent. We repeat this analysis across other models and observe the same trend, further demonstrating the generalizability of FAMA. We also find that, in some cases, using only a subset of the agents recommended by the mitigation agent can yield competitive or even improved performance. Overall, these results highlight the effectiveness of the FAMA framework in selecting appropriate agent combinations.

\begin{figure}[h]
    \centering
    \includegraphics[width=0.92\linewidth]{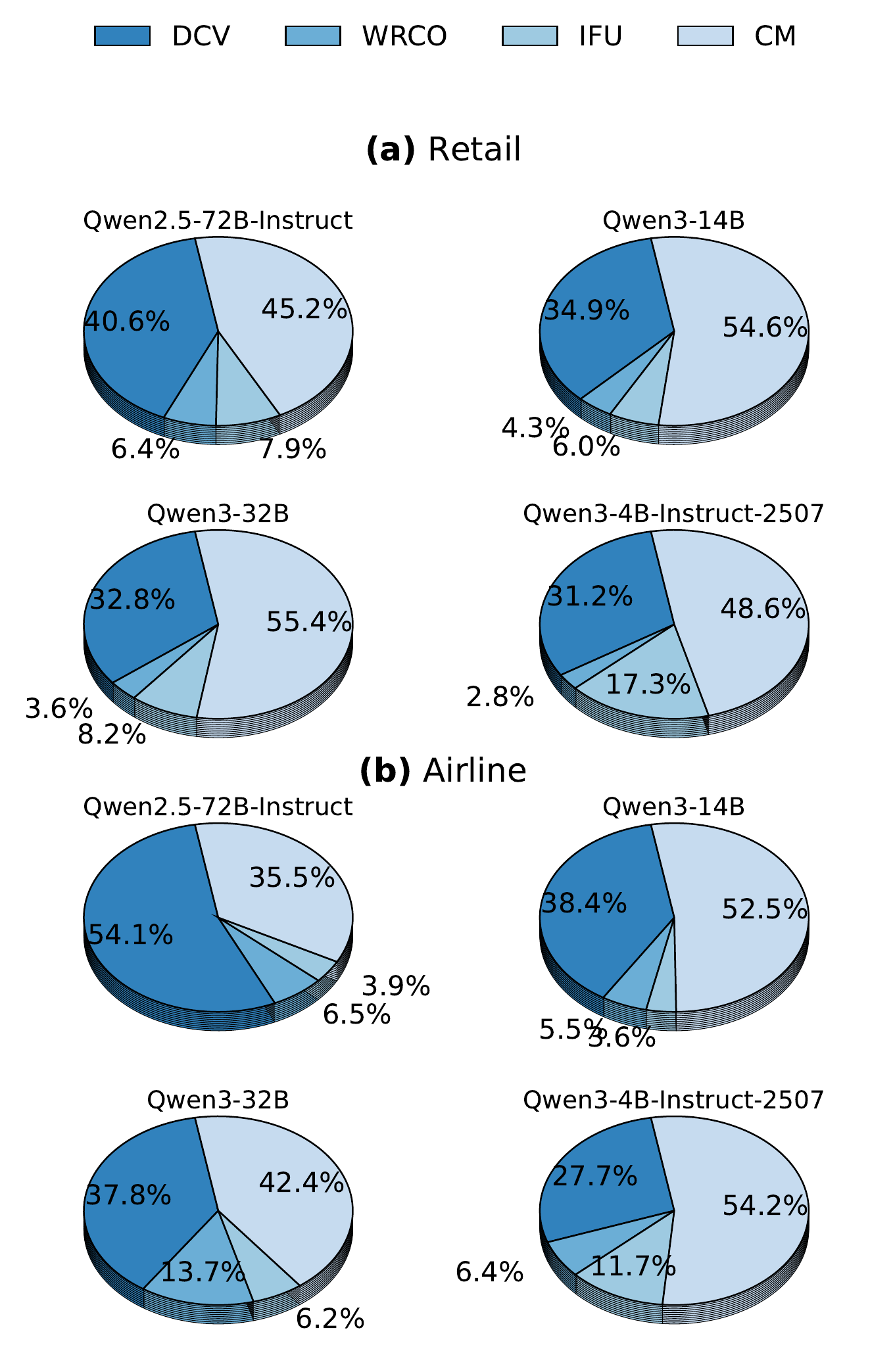}
    \caption{Statistical comparison of error categories across open-source LLMs in $\tau$-bench, as identified by the Orchestrator Agent using GPT-4.1-mini.}
    \label{fig:error_cat_gpt_4_1_mini}
\end{figure}

\begin{figure}[h]
    \centering
    \includegraphics[width=1\linewidth]{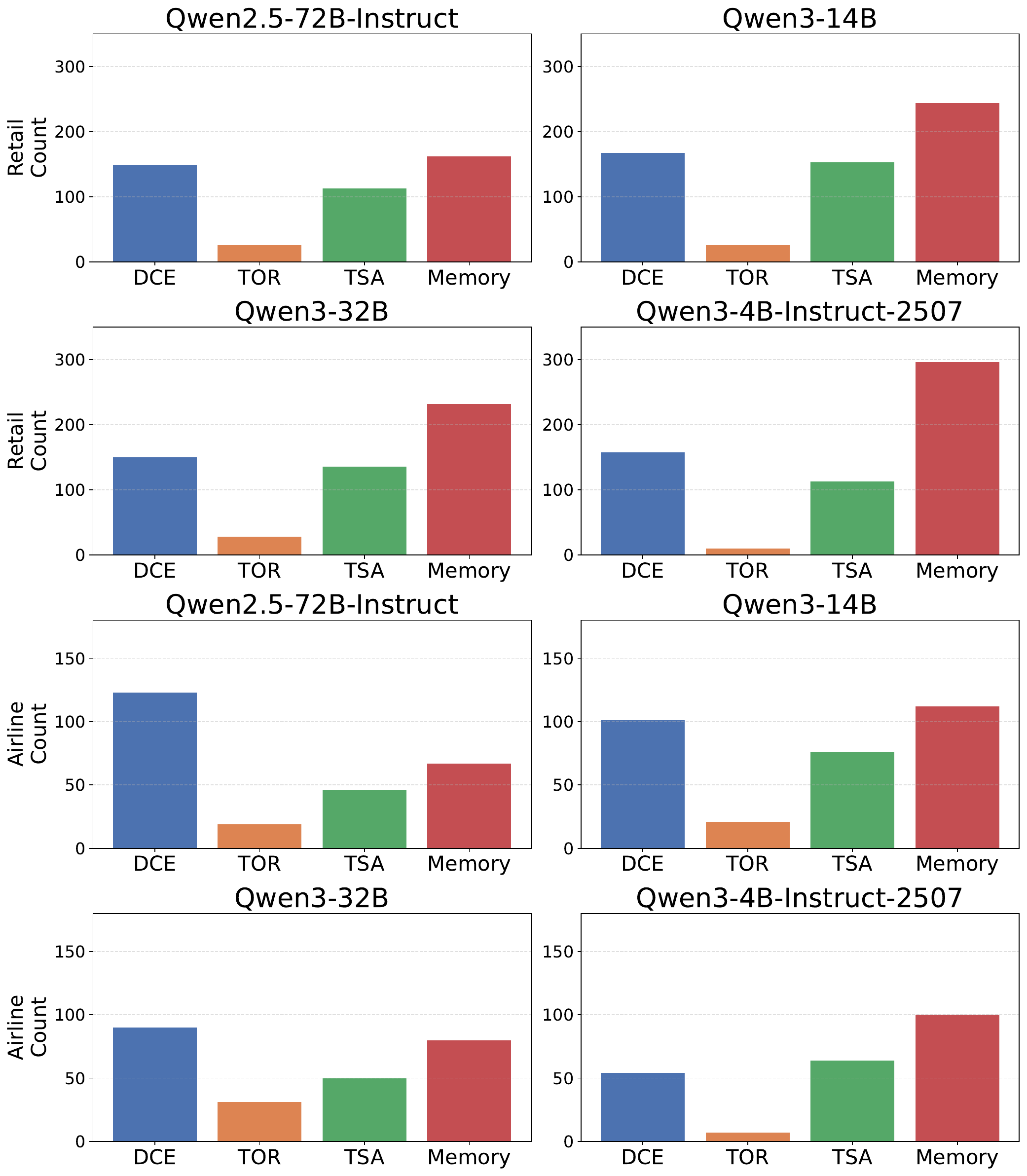}
    \caption{Statistics of recommendations produced by the Mitigation Agent using GPT-4.1-mini across different open source tool-use agents.}
    \label{fig:bar_chart_gpt_4_1_mini}
\end{figure}

\paragraph{Different judgment models produce the same decisions.} For failure analysis and agent selection within the FAMA framework, we initially use GPT-4o. To evaluate the robustness of the decision process, we repeat the same analysis and selection procedure on baseline results using GPT-4.1-mini as the judgment model. The results in Figure \ref{fig:error_cat_gpt_4_1_mini} show that, similar to GPT-4o, the alternative judgment model identifies Contextual Misinterpretation (CM) and Domain Constraint Violation (DCV) as the primary failure modes of open-source models. Accordingly, the mitigation agent consistently recommends the Memory module and the Domain Constraints Extractor agent, as shown in Figure \ref{fig:bar_chart_gpt_4_1_mini}.

\begin{figure*}
    \centering
    \includegraphics[width=1\linewidth]{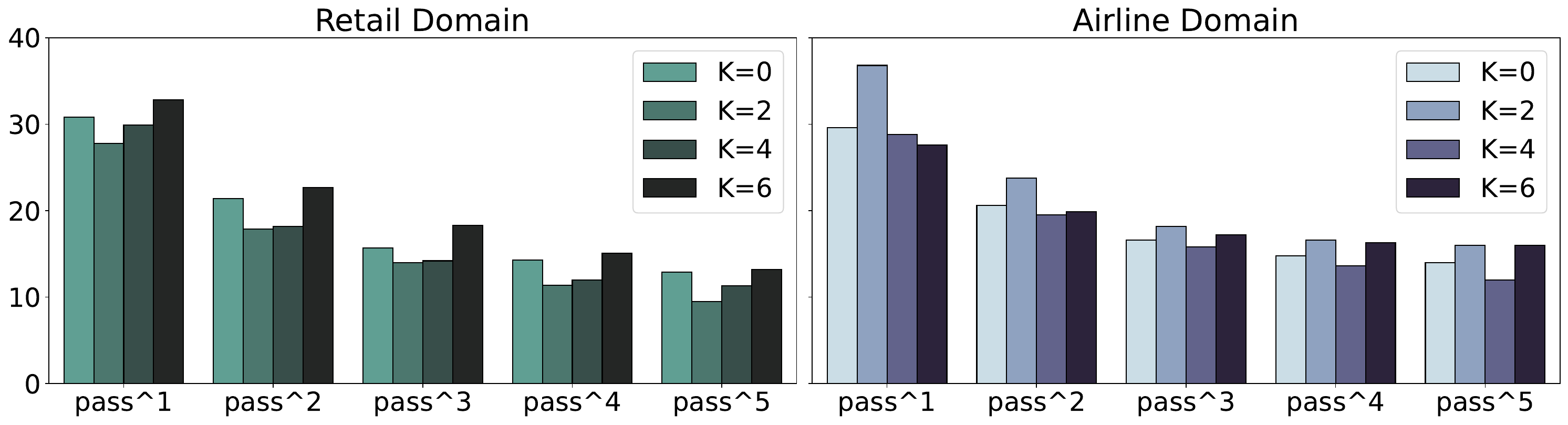}
    \caption{Comparison of agent performance in $\tau$-bench across different values of $k$, where $k$ denotes the number of most recent user queries retained.}
    \label{fig:ablation_mem_tau_bench}
\end{figure*}

These findings indicate that different frontier models exhibit consistent behavior in failure analysis and agent selection within the FAMA framework.

\section{Detail of Benchmarks}
\label{sec:appendix_benchmark_details}

\input{tables/main_res_ace_bench}

Each task in $\tau$-bench, $\tau$-trait and ACEBench is formulated as a Partially Observable Markov Decision Process (POMDP) \citep{ASTROM1965174}, in which an assistant agent interacts with a user entity/agent through a sequence of natural language conversations and tool invocations. At each step, the agent must decide which function calls to issue based on the current conversational context and partial observations derived from the external environment. Executed function calls produce structured outputs that are incorporated into the evolving dialogue and inform subsequent decisions. This interaction proceeds over multiple turns until task termination, at which point the agent is evaluated using the appropriate benchmark-prescribed metric centered around successful conversation/task completion.

\paragraph{$\tau$-bench}
$\tau$-bench evaluates tool using conversational agents in two real-world-style customer-service domains—retail and airline, where an agent must interact with a simulated user, follow domain policy rules, and use API tools to update/query a realistic backend state. In retail, the benchmark includes a database with 500 users, 50 products, and 1,000 orders, exposes 15 tools total (7 write + 8 non-write), and provides 115 tasks spanning actions like cancellations, returns/exchanges, address changes, and information requests; in airline, it includes 500 users, 300 flights, and 2,000 reservations, has 13 tools (6 write + 7 non-write), and contains 50 tasks around booking/modifying/canceling reservations and refund-related workflows.

\paragraph{$\tau$-trait}
$\tau$-trait is a robustness-focused extension of $\tau$-bench that keeps the multi-turn tool-use setting but adds persona-aware user simulation and expands domain coverage beyond retail/airline to include telecom and telehealth, targeting robustness, personalization, and fairness under realistic behavior shifts. It contains 218 total tasks split across 4 distinct domains, like retail (120), airline (60), telecom (18), and telehealth (20).

For $\tau$-bench and $\tau$-trait, we evaluate the task success performance using the pass\textasciicircum k metric \citep{yao2024tau}. It is defined as the probability that all of the k independently sampled outputs successfully complete the task, averaged across all tasks. Specifically, if a task is run for $n$ independent trials and $c$ of those are successful (i.e., have a correct result with reward $r=1$), then pass\textasciicircum k can be computed using the following formula:

\begin{equation*}
    \text{pass\textasciicircum k} = \mathbb{E}_{\text{task}}\left[ \binom{c}{k} \middle/ \binom{n}{k} \right]
\end{equation*}

\paragraph{ACEBench}
ACEBench is a tool used for benchmarking centered on function-call correctness across broad everyday domains, built in two parallel languages (English and Chinese) with a final size of 2,000 annotated entries. It includes a large synthetic API collection with 4,538 APIs spanning 8 major domains and 68 sub-domains (including technology, finance, entertainment, society, health, culture, environment, and others). From 50 agent-based tasks, we have mainly focused on 30 agent multi-turn tasks for our work, discarding the other 20 multi-step tasks. 

For ACEBench, End-to-End Accuracy measures the degree of correspondence between
the predicted instance attributes and the ground truth. When all attributes align perfectly, the accuracy equals 1; any discrepancy results in an
accuracy of 0. Process Accuracy quantifies how closely the actual sequence of function calls aligns with the expected (ideal) sequence. This metric is defined as $\frac{n}{m}$, where $m$ denotes the total number of steps in the ideal function call sequence, and $n$ indicates the number of steps where the actual process matches the ideal one.

\section{Analysis of the $\tau$-trait Results}
\label{appendix:tau_trait_results}
To demonstrate the generalizability of the FAMA framework, we evaluate it on $\tau$-trait, which includes four domains: Retail, Airline, Telehealth, and Telecom. Since we have already analyzed performance on the Airline and Retail domains in $\tau$-bench, we focus our evaluation on the Telehealth and Telecom domains.

\begin{figure*}
    \centering
    \includegraphics[width=1\linewidth]{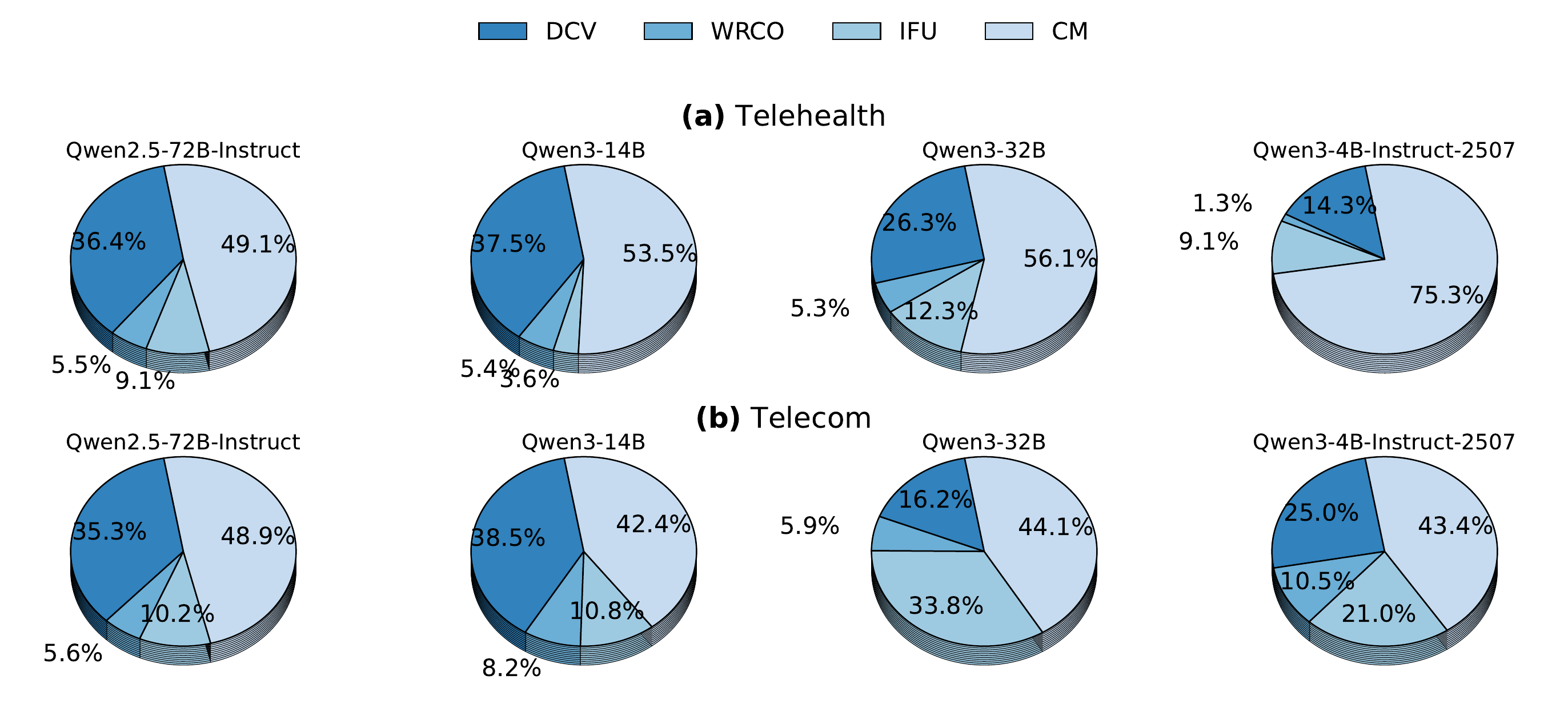}
    \caption{Statistical comparison of error categories across open-source LLMs in $\tau$-trait environment, as identified by the Orchestrator Agent.}
    \label{fig:orch_error_category_tau_trait}
\end{figure*}

\begin{figure*}
    \centering
    \includegraphics[width=1\linewidth]{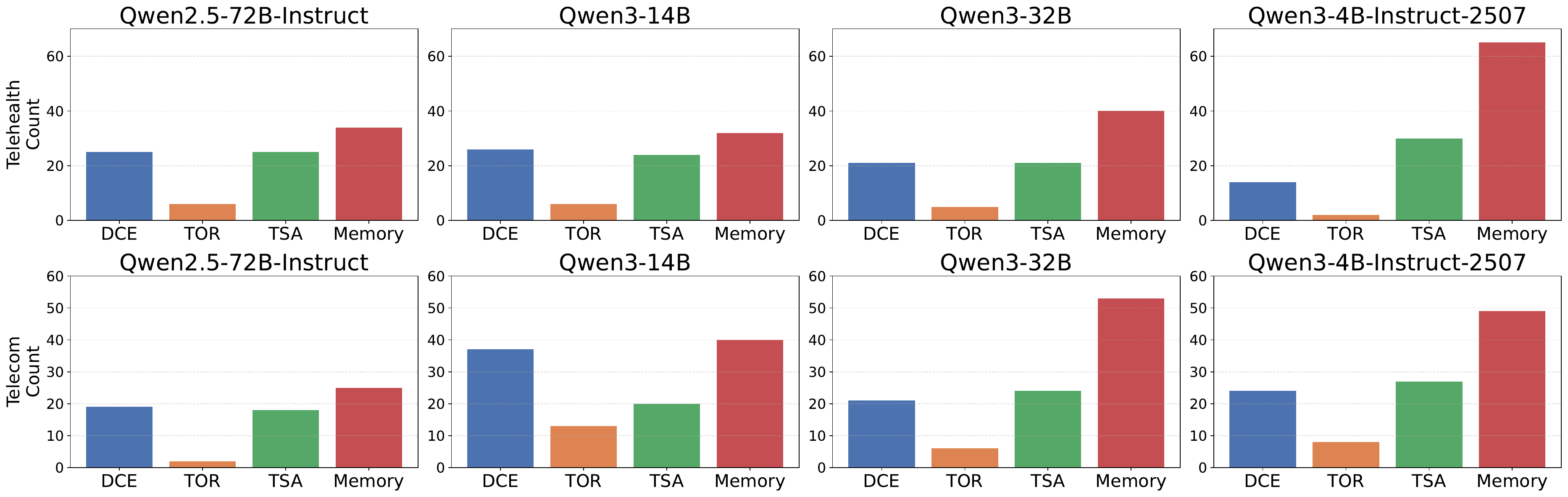}
    \caption{Statistics of recommendations produced by the Mitigation Agent across different open source agents.}
    \label{fig:mitigation_agent_tau_trait}
\end{figure*}

Similar to $\tau$-bench, Domain Constraint Violations (DCV) and Contextual Misinterpretation (CM) are identified as critical error sources by the orchestrator agent, as shown in Figure \ref{fig:orch_error_category_tau_trait}. However, unlike Tau-Bench, memory-related issues are more prominent across Tau-Trait domains, with Incomplete Fulfillment (IFU) emerging as the primary cause of task failures. As illustrated in Figure \ref{fig:mitigation_agent_tau_trait}, the mitigation agent frequently recommends the Memory module for a larger number of models. Consequently, using only the Memory module significantly improves the performance of open-source models and, in some cases, yields better results than combinations that also include the Domain Constraints Extractor (DCE) agent.

The results in Table~\ref{tab:tau_trait_results} show that FAMA consistently outperforms IRMA across both the Telecom and Telehealth domains on \text{pass\textasciicircum k}, for $\text{k} = 1, \ldots, 5$. While certain methods, such as FC, outperform FAMA in isolated cases, FC does not demonstrate consistent performance across domains and exhibits substantial variability depending on the domain and model. These results further highlight the robustness and generalizability of the FAMA framework across diverse settings.

\input{tables/main_res_tau_trait}

\section{Analysis of the ACEBench Results}
\label{appendix:ace_bench_results}
ACEBench is another benchmark used to evaluate the FAMA framework. Although ACEBench supports multiple evaluation settings, we focus on the multi-turn agent configuration, where an active user agent interacts with a tool-calling agent. We compare FAMA against the Base Agent defined in ACEBench and the ReAct framework.

\begin{figure*}
    \centering
    \includegraphics[width=1\linewidth]{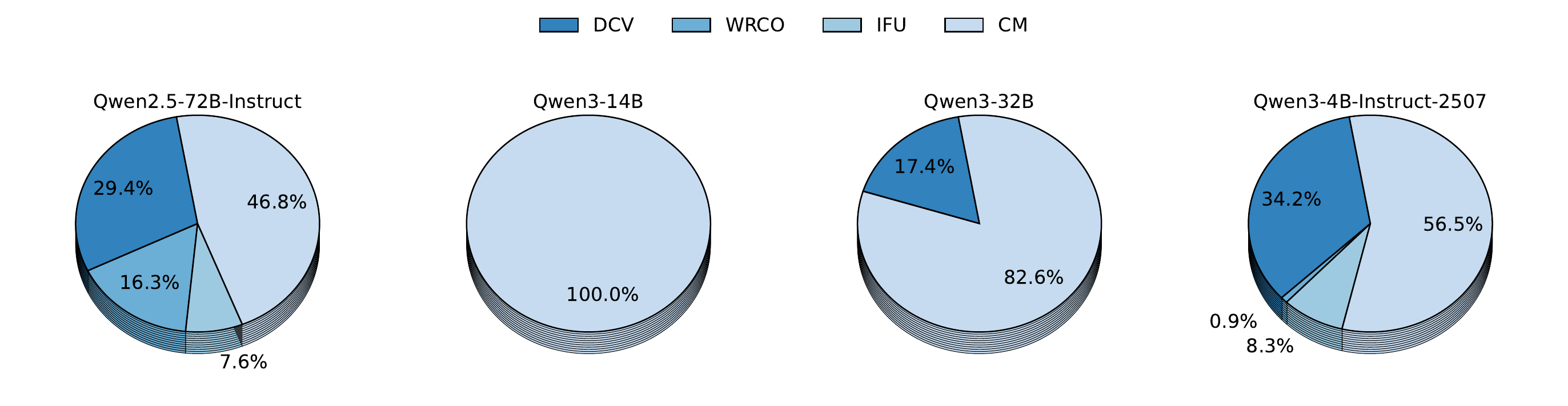}
    \caption{Statistical comparison of error categories across open-source LLMs in ACEBench, as identified by the Orchestrator Agent.}
    \label{fig:orch_error_category_ace_bench}
\end{figure*}

\begin{figure*}
    \centering
    \includegraphics[width=1\linewidth]{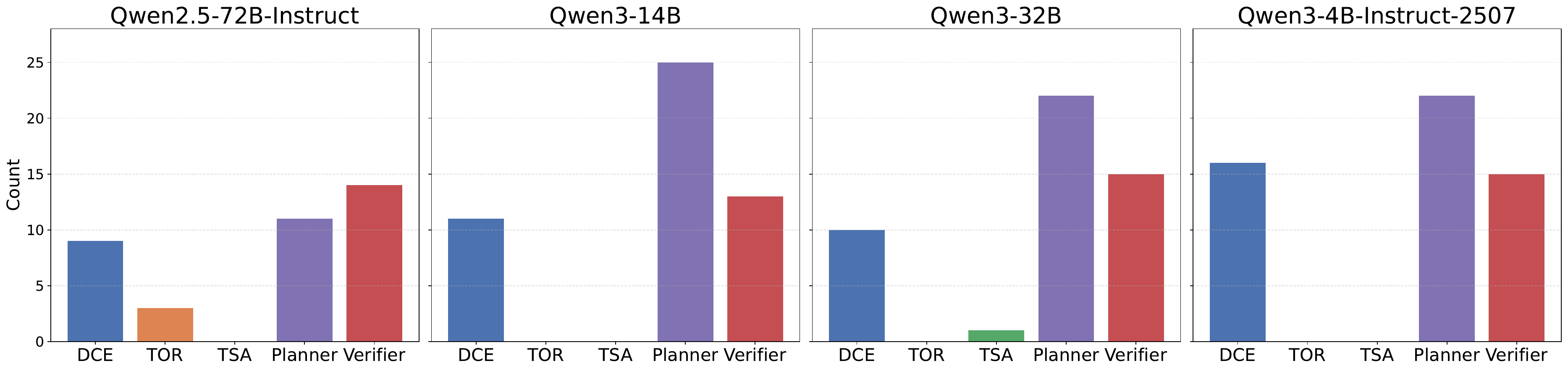}
    \caption{Statistics of recommendations produced by the Mitigation Agent across different open source agents for ACEBench.}
    \label{fig:agent_suggestion_mitigation_agent_ace_bench}
\end{figure*}

The results in Figure \ref{fig:orch_error_category_ace_bench} illustrate the distribution of failure causes on ACEBench as identified by the orchestrator agent. As shown in the figure, all evaluated models suffer from significant Contextual Misinterpretation (CM) issues, manifested as incomplete task execution, misinterpretation during problem solving, and incorrect interactions with the environment. To address these issues, the mitigation agent consistently recommends the inclusion of Planner and Verifier agents, which are well-suited to correcting such failures as shown in Figure \ref{fig:agent_suggestion_mitigation_agent_ace_bench}.

The results reported in Table \ref{tab:ace_bench_results} show that augmenting ReAct and the Base Agent with FAMA leads to consistent performance improvements over other baselines, even with the addition of only two auxiliary agents. In the following section, we present a qualitative example illustrating how FAMA mitigates these issues and successfully completes a task.

\subsection{Example}
In this example, we use Qwen2.5-72B-Instruct as both the tool-calling agent and the user agent. The example corresponds to a message-sending task from ACE-Bench. We compare the interaction produced by the Base Agent, as defined in ACEBench, with the interaction generated using the FAMA framework.

\begin{tcolorbox}[colback=white, colframe=black, fonttitle=\bfseries, title=Example: Baseline, breakable]
\small

\textbf{User:} Can you help me send a message to Frank? The content is: `Are you free tomorrow to have lunch with me?'

\textbf{Assistant:} send\_message(sender\_name='user', receiver\_name='Frank', message='Are you free tomorrow to have lunch with me?')

\textbf{Tool:} ``status'': false, ``message'': ``Wi-Fi is turned off, cannot send messages at this time''

\textbf{Assistant:} turn\_on\_wifi()

\textbf{Tool:} ``status'': true, ``message'': ``Wi-Fi has been turned on''

\textbf{Assistant:} send\_message(sender\_name=`user', receiver\_name=`Frank', message=`Are you free tomorrow to have lunch with me?')

\textbf{Tool:} ``status'': false, ``message'': ``Inbox capacity is full. You need to ask the user which message to delete.

\textbf{Assistant:} Memory is full, which message would you like to delete?

\textbf{User:} Please delete the oldest message.

\textbf{Assistant:} get\_earliest\_message\_id()

\textbf{Tool:} ``status'': true, ``message'': ``The earliest message ID is 3'', ``message\_id'': 3

\textbf{Assistant:} delete\_message(message\_id=3)

\textbf{Tool:} ``status'': true, ``message'': ``Message ID 3 has been successfully deleted.''

\textbf{Assistant:} send\_message(sender\_name=`user', receiver\_name=`Frank', message=`Are you free tomorrow to have lunch with me?')

\textbf{Tool:} ``status'': false, ``message'': ``Sender or receiver does not exist''

\textbf{Assistant:} Could you please confirm the correct name of the recipient?

\textbf{User:} The recipient's name is indeed Frank. Could you proceed with sending the message?

\textbf{Assistant:} send\_message(...) $\rightarrow$ \textcolor{red}{Repeated failures...}

\textbf{User:} The recipient's name is correct... Here is his phone number: 123-456-7890.

\textbf{Assistant:} send\_message(..., phone\_number=`123-456-7890')

\textbf{User:} Here is his email address frank@example.com

\textbf{Assistant:} send\_message(email=`frank@example.com') 

\vspace{1mm}
\textbf{Result:} \textcolor{red}{Task Failed} -- Agent loops without resolving the sender name issue.

\end{tcolorbox}

\vspace{3mm}

\begin{tcolorbox}[colback=white, colframe=black, fonttitle=\bfseries, title=FAMA Framework, breakable]
\small

\textbf{User:} Can you help me send a message to Frank? The content is: `Are you free tomorrow to have lunch with me?'

\textbf{Planner:} \\1. Ensure device is logged in: login\_device()\\
2. Send message to Frank\\
3. Confirm with user about reminder\\
4. Finish conversation

\textbf{Assistant:} login\_device()

\textbf{Verifier:} \textcolor{green!50!black}{PASS}; Correctly initiated device login.

\textbf{Tool:} ``status'': true, ``message'': ``Device has been logged in''

\textbf{Assistant:} send\_message(sender\_name=`User', receiver\_name=`Frank', message=`...')

\textbf{Verifier:} \textcolor{green!50!black}{PASS}; Correctly sent the message as per the plan.

\textbf{Tool:} ``status'': false, ``message'': ``Wi-Fi is turned off...''

\textbf{Assistant:} turn\_on\_wifi()

\textbf{Verifier:} \textcolor{green!50!black}{PASS}; Correctly identified the need to turn on Wi-Fi.

\textbf{Tool:} ``status'': true, ``message'': ``Wi-Fi has been turned on''

\textbf{Assistant:} send\_message(...)

\textbf{Verifier:} \textcolor{orange}{RECHECK}; Wi-Fi was turned on but message was not sent; fix: retry send\_message

\textbf{Tool:} ``status'': false, ``message'': ``Inbox capacity is full...''

\textbf{Assistant:} Memory is full, which message would you like to delete?

\textbf{Verifier:} \textcolor{green!50!black}{PASS}; Correctly identified the need to delete a message.

\textbf{User:} Please delete the oldest message.

\textbf{Planner:} \\1. get\_earliest\_message\_id()\\
2. delete\_message(message\_id=EARLIEST MESSAGE ID)\\
3. Confirm deletion

\textbf{Assistant:} get\_earliest\_message\_id()

\textbf{Verifier:} \textcolor{green!50!black}{PASS}; Correctly identified need to get earliest message ID.

\textbf{Tool:} message\_id: 3

\textbf{Assistant:} delete\_message(message\_id=3)

\textbf{Verifier:} \textcolor{green!50!black}{PASS}; Correctly deletes the oldest message as requested.

\textbf{Tool:} `status'': true, ``message'': ``Message ID 3 deleted.''

\textbf{User:} Sure, let's try again.

\textbf{Assistant:} send\_message(sender\_name=`User', receiver\_name=`Frank', ...)

\textbf{Verifier:} \textcolor{green!50!black}{PASS}; Correctly attempting to send message after clearing memory.

\textbf{Tool:} ``status'': false, ``message'': ``Sender or receiver does not exist''

\textbf{Assistant:} Could you please confirm the correct names for sender and receiver?

\textbf{Verifier:} \textcolor{orange}{RECHECK}; Sender or receiver does not exist; fix: Ask user to confirm names.

\textbf{User:} The sender is Grace, and the receiver is Frank.

\textbf{Planner:} 1. Ensure device is logged in\\
2. Send message from Grace to Frank

\textbf{Assistant:} send\_message(sender\_name=`Grace', receiver\_name=`Frank', message=`Are you free
tomorrow to have lunch with me?')

\textbf{Verifier:} \textcolor{green!50!black}{PASS}; Correctly updated sender name and resent message.

\textbf{Tool:} ``status'': true, ``message'': ``Message successfully sent to Frank.''

\textbf{Assistant:} The message has been successfully sent to Frank. Is there anything else you need help with?

\textbf{Verifier:} \textcolor{green!50!black}{PASS}; Correctly sent the message and confirmed success.

\textbf{User:} finish conversation.

\textbf{Planner:}
\\1. Ask the user if they have any final tasks or requests before finishing the conversation.\\2. If the user has no further tasks or requests, respond with \"finish conversation\" to end the dialogue.

\vspace{1mm}
\textbf{Result:} \textcolor{green!50!black}{Task Completed Successfully with Full Verification}

\end{tcolorbox}

The above example demonstrates that incorporating only a Planner and a Verifier agent, without additional auxiliary agents, is sufficient to successfully solve a tool-calling task.

\section{Error Categories and Causes of the Failures}
\label{appendix:err_cat_def}
In this section, we first describe the error categories and then elaborate on the underlying causes of failure associated with each category. The definition of each error category is provided below.

\paragraph{Domain Policy Violation:} This error refers to a failure mode in which a tool-using agent issues an action or tool invocation that contravenes the explicit operational rules, constraints, or policies of the simulated domain defined by the environment. These policies govern which actions are permissible under specific states and preconditions. A violation occurs when the agent ignores, misinterprets, or incorrectly applies these domain rules, such as invoking a return operation that is disallowed by the retail domain’s return policy in $\tau$-bench, resulting in an invalid state transition. Because the environment enforces domain policies strictly, such violations cause the task execution trajectory to terminate unsuccessfully, even if the agent’s intent aligns with the high-level task objective. 

\paragraph{Incorrect Retrieval from Complex Tool Outputs:} refers to a failure mode in which a tool-using agent fails to correctly extract relevant information from prior tool responses stored in its context. This error typically arises when tool outputs are large, unstructured, or weakly structured (e.g., verbose logs, nested text fields, or mixed natural language and metadata), making it difficult for the agent to reliably identify the correct entities, values, or state variables required for subsequent actions. As a result, the agent conditions later tool calls or reasoning steps on incorrect or incomplete information, leading to invalid decisions or task failure despite the necessary information being present in the environment output.

\paragraph{Contextual Misinterpretation and Hallucination:} refers to a failure mode in which a tool-using agent incorrectly infers the user’s underlying intent due to nuanced, ambiguous, or adversarial natural language instructions. Instead of grounding its interpretation in the domain’s available operations and constraints, the agent over-literalizes phrasing, overlooks implied intent, or hallucinates an incorrect task formulation. This results in the selection and invocation of an inappropriate tool, such as executing a return operation when the user implicitly intends an exchange, despite the correct intent being inferable from contextual or domain knowledge.

\paragraph{Incomplete Fulfillment or Early Stopping:} refers to a failure mode in which a tool-using agent does not fully satisfy all components of the user’s task requirements before terminating or effectively abandoning the interaction. This error commonly arises in long-horizon, multi-turn trajectories where the agent must track multiple obligations across conversational turns while operating over large, unstructured tool outputs. As the interaction progresses, the agent may lose track of pending subgoals, become over-focused on intermediate tool calls, or prematurely conclude the task, resulting in unmet user requests despite partial progress.

To enable judgment agents to detect task failures, a detailed list of failure causes is required. Accordingly, we analyze the trajectories produced by baseline methods and provide a comprehensive list of failure causes for each error category, as described below.

\begin{tcolorbox}[colback=white, colframe=black, fonttitle=\bfseries, title=Cause of Domain Policy Violation, breakable]
\small

\textbf{1.} The assistant performs an action that the policy explicitly forbids

\textbf{2.} The assistant skips a step that the policy explicitly requires

\textbf{3.} The assistant proceeds without meeting conditions the policy requires

\textbf{4.} The assistant makes commitments that contradict policy rules

\textbf{5.} The assistant uses data/arguments not provided by user or tools (fabrication)

\end{tcolorbox}

\vspace{3mm}

\begin{tcolorbox}[colback=white, colframe=black, fonttitle=\bfseries, title=Cause of Wrong Retrieval from Complex Tool Outputs, breakable]
\small

\textbf{1.} The agent states information that differs from what the tool output actually contains

\textbf{2.} The agent miscounts items or provides wrong totals from tool outputs

\textbf{3.} The agent selects wrong items when correct options exist in tool output

\textbf{4.} The agent misses relevant data in tool output that would change the action

\textbf{5.} The agent attributes data to the wrong entity when multiple exist in output

\textbf{6.} The agent uses wrong values (IDs, prices, dates, etc.) from tool output

\end{tcolorbox}

\vspace{3mm}

\begin{tcolorbox}[colback=white, colframe=black, fonttitle=\bfseries, title=Cause of Incomplete Fulfillment or Early Stopping , breakable]
\small
The user may have multiple requirements, questions, or actions they need completed. This error occurs when the agent:

\textbf{1.} Addresses only some parts of a multi-part request

\textbf{2.} Stops after encountering a difficulty without trying alternatives

\textbf{3.} Receives confirmation to proceed but never executes the action

\textbf{4.} Terminates the conversation with unfinished tasks

\end{tcolorbox}

\vspace{3mm}

\begin{tcolorbox}[colback=white, colframe=black, fonttitle=\bfseries, title=Cause of Contextual Misinterpretation and Hallucination, breakable]
\small

The user's request may have specific intent, context, or nuances that the agent must correctly understand to take appropriate actions. This error occurs when the agent:

\textbf{1.} Understands the words but misses the actual intent

\textbf{2.} Confuses which entity/item/order the user is referring to

\textbf{3.} Misapplies criteria or constraints the user specified

\textbf{4.} Takes an action that doesn't match what the user actually wanted

\end{tcolorbox}

\vspace{3mm}

\input{tables/best_comp_fama_tau_bench}

\input{tables/best_comp_fama_tau_trait}

\section{Optimal Components of FAMA}
\label{sec:optimal_fama_components}
In this section, we describe the auxiliary agents and modules commonly used in static multi-agent frameworks and present, for each model, the final set of components selected by FAMA that achieves the best performance reported in Tables \ref{tab:tau_bench_results}, \ref{tab:tau_trait_results}, and \ref{tab:ace_bench_results}.

For $\tau$-bench and $\tau$-trait, we assume the availability of the Memory module, the Domain Constraints Extractor (DCE) Agent, the Tool Output Reformulation (TOR) Agent, and the Tool Suggestion (TS) Agent. For ACE-Bench, we additionally include a Planner Agent and a Verifier Agent, as the base agent in ACE-Bench is unable to generate free-form text and can only invoke tools with their input arguments. The final set of agents used in the FAMA framework is presented in Tables \ref{tab:best_comp_tau_bench}, \ref{tab:best_comp_tau_trait}, and \ref{tab:best_comp_ace_bench}.

\input{tables/best_comp_fama_ace_bench}


\section{Comparison with Self-Reflection}
\label{appendix:self_reflection}

As part of the rebuttal, we also implement and compare against a self-reflection (SR) baseline, inspired by approaches such as Self-RAG and “System 2”-style prompting \citep{madaan2023self, shinn2023reflexion}. This baseline augments the base tool-use agent with an explicit reflection and correction phase, where intermediate outputs are analyzed and iteratively revised before producing the final response.

\input{tables/thinking_model_error_frequency}

We evaluate this SR baseline alongside FAMA and ReAct across multiple models in $\tau$-bench. Our results (Tables \ref{tab:SR_comparisons_1}-\ref{tab:SR_comparisons_4}) show that FAMA consistently outperforms the SR baseline, suggesting that structured failure analysis combined with targeted mitigation is more effective than relying solely on generic reflection and revision mechanisms in our setting.

\label{sec:Self_Reflection}
\input{tables/SR_qwen72B}
\input{tables/SR_qwen32B}
\input{tables/SR_qwen14B}
\input{tables/SR_qwen4B}

%% file: tables/token_latency_analysis_tau_bench.tex

\begin{table*}[t]
\centering
\small
\begin{tabular}{lrrrrrrrr}
\toprule
\multicolumn{8}{c}{\textbf{$\tau$-bench-airline}} \\
\midrule
\textbf{Method} & \textbf{Min} & \textbf{Max} & \textbf{Median} & \textbf{Avg} & \textbf{Assistant} & \textbf{Overhead} & \textbf{Overhead} & \textbf{Latency} \\
 & \textbf{Tokens} & \textbf{Tokens} & \textbf{Tokens} & \textbf{Tokens} & \textbf{Tokens} & \textbf{Tokens} & \textbf{(\%)} & \textbf{(seconds)}\\
\midrule
React-non-thinking & 96 & 4591 & 971.5 & 1148.9 & 1148.9 & 0 & 0 & 59.99 \\
React-thinking	& 269 & 32175 & 	4957.5 & 	5729.6 & 	5729.6 & 	0 & 	0 & 	221.43\\
IRMA  & 346 & 14324 & 2725 & 3511.1 & 1694.4 & 1816.7 & 50 & 149.78 \\
FAMA  & 358 & 8111 & 2137 & 2617.4 & 1822.1 & 795.3 & 30 & 91.12 \\
\midrule
\multicolumn{8}{c}{\textbf{$\tau$-bench-retail}} \\
\midrule
\textbf{Method} & \textbf{Min} & \textbf{Max} & \textbf{Median} & \textbf{Avg} & \textbf{Assistant} & \textbf{Overhead} & \textbf{Overhead} & \textbf{Latency} \\
 & \textbf{Tokens} & \textbf{Tokens} & \textbf{Tokens} & \textbf{Tokens} & \textbf{Tokens} & \textbf{Tokens} & \textbf{(\%)} & \textbf{(seconds)} \\
\midrule
React-non-thinking & 59 & 3727 & 1108 & 1131.4 & 1131.4 & 0 & 0 & 49.05 \\
React-thinking	& 246	& 32768	& 4331	& 4623.4 &	4623.4	& 0 &	0 &	189.79\\
IRMA  & 380 & 32768 & 3421 & 3980.4 & 1671.72 & 2308.7 & 58 & 111.62 \\
FAMA  & 161 & 8040 & 2167 & 2440 & 1714.6 & 725.4 & 29.7 & 56.73 \\
\bottomrule
\end{tabular}
\caption{Token and latency cost analysis of various methods for $\tau$-Airline and $\tau$-Retail. The tokens reported in Columns 2–5 refer to the sum of (i) tokens generated by the main tool-use assistant (Column 6) and (ii) additional overhead tokens introduced by helper agents (Column 7). “Assistant tokens” denote the average tokens produced solely by the primary tool-calling agent, while “overhead tokens” correspond to the average tokens generated by helper agents in the IRMA and FAMA settings. `ReAct-non-thinking’ and `ReAct-thinking’ represent non-thinking and thinking models used for ReAct method.}
\label{tab:token_latency_overhead_analysis}
\end{table*}

%% file: tables/main_res_ace_bench.tex
\begin{table}[t]
\centering
\small
\begin{tabular}{lcc}
\toprule
\textbf{Method} & \textbf{End-to-End Acc} & \textbf{Process Acc} \\
\midrule

\multicolumn{3}{l}{\textit{\textbf{Qwen3-4B-Instruct}}} \\
\addlinespace[3pt]
\hspace{1em}Base        & 10.0\%   & 41.6\% \\
\hspace{1em}ReAct       & 10.0\%   & \underline{51.3}\% \\
\hspace{1em}FAMA-Base (\textbf{our})  & \underline{16.7}\% & 45.1\% \\
\hspace{1em}FAMA-ReAct (\textbf{our}) & \textbf{20.0}\%   & \textbf{53.9}\% \\
\addlinespace[3pt]
\hdashline
\addlinespace[3pt]
\multicolumn{3}{l}{\textit{\textbf{Qwen3-14B}}} \\
\addlinespace[3pt]
\hspace{1em}Base        & 16.7\% & 40.8\% \\
\hspace{1em}ReAct       & \underline{20.0}\%   & \textbf{61.3}\% \\
\hspace{1em}FAMA-Base (\textbf{our})  & \textbf{23.3}\% & 50.0\%   \\
\hspace{1em}FAMA-ReAct (\textbf{our})  & \textbf{23.3}\% & \underline{58.4}\% \\
\addlinespace[3pt]
\hdashline
\addlinespace[3pt]
\multicolumn{3}{l}{\textit{\textbf{Qwen3-32B}}} \\
\addlinespace[3pt]
\hspace{1em}Base        & \underline{23.3}\% & 46.5\% \\
\hspace{1em}ReAct       & 13.3\% & \underline{50.2}\% \\
\hspace{1em}FAMA-Base (\textbf{our})   & \underline{23.3}\% & 50.0\%   \\
\hspace{1em}FAMA-ReAct (\textbf{our})  & \textbf{30.0}\%   & \textbf{59.6}\% \\
\addlinespace[3pt]
\hdashline
\addlinespace[3pt]
\multicolumn{3}{l}{\textit{\textbf{Qwen2.5-72B-Instruct}}} \\
\addlinespace[3pt]
\hspace{1em}Base        & 23.3\% & 45.1\% \\
\hspace{1em}ReAct       & 36.7\% & 57.6\% \\
\hspace{1em}FAMA-Base (\textbf{our})   & \textbf{50.0}\%   & \textbf{70.2}\% \\
\hspace{1em}FAMA-ReAct (\textbf{our}) & \underline{43.3}\% & \underline{65.2}\% \\

\bottomrule
\end{tabular}
\caption{Comparison of agentic approaches across different open-source agents on ACEBench.}
\label{tab:ace_bench_results}
\end{table}

%% file: tables/main_res_tau_trait.tex
\begin{table*}[t]
\centering
\small
\begin{tabular}{lrrrrrrrrrr}
\toprule
\textbf{Method} & \multicolumn{5}{c}{\textbf{$\tau$-Telehealth}} & \multicolumn{5}{c}{\textbf{$\tau$-Telecom}} \\
\cmidrule(lr){2-6}\cmidrule(lr){7-11}
& \multicolumn{1}{l}{\textbf{Pass\textasciicircum{}1}}
& \multicolumn{1}{l}{\textbf{Pass\textasciicircum{}2}}
& \multicolumn{1}{l}{\textbf{Pass\textasciicircum{}3}}
& \multicolumn{1}{l}{\textbf{Pass\textasciicircum{}4}}
& \multicolumn{1}{l}{\textbf{Pass\textasciicircum{}5}}
& \multicolumn{1}{l}{\textbf{Pass\textasciicircum{}1}}
& \multicolumn{1}{l}{\textbf{Pass\textasciicircum{}2}}
& \multicolumn{1}{l}{\textbf{Pass\textasciicircum{}3}}
& \multicolumn{1}{l}{\textbf{Pass\textasciicircum{}4}}
& \multicolumn{1}{l}{\textbf{Pass\textasciicircum{}5}} \\
\midrule

\multicolumn{11}{l}{\textbf{\textit{Qwen3-4B-Instruct}}} \\
\addlinespace[3pt]
ReAct & 22.99\% & 16.00\% & 12.50\% & 11.00\% & 10.00\%
      & 15.55\% & 8.33\% & 6.11\% & 5.55\% & 5.55\% \\
FC    & \textbf{34.00}\% & \textbf{27.00}\% & \textbf{24.00}\% & \textbf{22.00}\% & \textbf{20.00}\%
      & 37.00\% & \textbf{31.00}\% & \textbf{27.00}\% & \textbf{24.00}\% & \textbf{21.00}\% \\
IRMA  & 22.99\% & 14.50\% & 11.00\% & 10.00\% & 10.00\%
      & \underline{38.00}\% & 26.00\% & 21.00\% & \underline{18.00}\% & 16.00\% \\
FAMA (\textbf{our})  & \underline{32.00}\% & \underline{24.00}\% & \underline{20.00}\% & \underline{18.00}\% & \underline{17.00}\%
      & \textbf{39.00}\% & \underline{28.00}\% & \underline{22.00}\% & \underline{18.00}\% & \underline{17.00}\% \\
\addlinespace[3pt]
\hdashline

\addlinespace[3pt]
\multicolumn{11}{l}{\textit{\textbf{Qwen3-14B}}} \\
\addlinespace[3pt]
ReAct & \textbf{44.00}\% & \textbf{43.00}\% & \textbf{42.00}\% & \textbf{41.00}\% & \textbf{40.00}\%
      & 15.55\% & 11.11\% & 11.11\% & 11.11\% & 11.11\% \\
FC    & 39.00\% & 38.00\% & 37.00\% & 36.00\% & 35.00\%
      & \underline{38.00}\% & \underline{33.33}\% & \underline{30.00}\% & \underline{28.00}\% & \underline{27.00}\% \\
IRMA  & 39.99\% & 27.99\% & 24.50\% & 22.00\% & 20.00\%
      & 36.66\% & 23.30\% & 16.66\% & 13.33\% & 11.11\% \\
FAMA (\textbf{our}) & \underline{42.00}\% & \underline{41.00}\% & \underline{40.00}\% & \underline{40.00}\% & \textbf{40.00}\%
      & \textbf{43.00}\% & \textbf{35.50}\% & \textbf{31.00}\% & \textbf{29.00}\% & \textbf{28.00}\% \\
\addlinespace[3pt]
\hdashline

\addlinespace[3pt]
\multicolumn{11}{l}{\textit{\textbf{Qwen3-32B}}} \\
\addlinespace[3pt]
ReAct & \underline{43.00}\% & \underline{37.00}\% & 34.00\% & 32.00\% & 30.00\%
      & 24.44\% & 20.00\% & 18.88\% & 17.77\% & 16.66\% \\
FC    & 42.00\% & \textbf{39.00}\% & \textbf{37.00}\% & \underline{35.00}\% & \textbf{35.00}\%
      & 32.00\% & 25.00\% & 21.00\% & 18.00\% & 16.00\% \\
IRMA  & 30.00\% & 24.50\% & 20.00\% & 18.00\% & 15.00\%
      & \textbf{40.00}\% & \underline{28.00}\% & \underline{24.44}\% & \underline{23.33}\% & \underline{22.20}\% \\
FAMA (\textbf{our})  & \textbf{45.00}\% & \underline{37.00}\% & \underline{35.00}\% & \textbf{35.00}\% & \textbf{35.00}\%
      & \underline{37.77}\% & \textbf{31.66}\% & \textbf{30.00}\% & \textbf{28.88}\% & \textbf{27.77}\% \\
\addlinespace[3pt]
\hdashline

\addlinespace[3pt]
\multicolumn{11}{l}{\textit{\textbf{Qwen2.5-72B-Instruct}}} \\
\addlinespace[3pt]
ReAct & \textbf{43.00}\% & \underline{39.00}\% & \underline{38.00}\% & \underline{35.00}\% & \underline{34.00}\%
      & \underline{51.11}\% & \underline{39.44}\% & \underline{32.22}\% & 26.66\% & 22.22\% \\
FC    & 41.00\% & 36.00\% & 32.00\% & 31.00\% & 30.00\%
      & 44.00\% & 35.00\% & 31.00\% & \underline{28.00}\% & \underline{27.00}\% \\
IRMA  & 31.00\% & 28.50\% & 27.00\% & 26.00\% & 25.00\%
      & 44.44\% & 30.00\% & 23.88\% & 22.22\% & 22.22\% \\
FAMA (\textbf{our})  & \underline{42.00}\% & \textbf{40.00}\% & \textbf{40.00}\% & \textbf{40.00}\% & \textbf{40.00}\%
      & \textbf{52.00}\% & \textbf{44.00}\% & \textbf{41.10}\% & \textbf{40.00}\% & \textbf{39.00}\% \\

\bottomrule
\end{tabular}
\caption{Comparison of different open-source agents on $\tau$-trait across five runs. Note that Qwen2.5-72B-Instruct is used as the model backbone of the sub-agents of IRMA and FAMA agentic frameworks.}
\label{tab:tau_trait_results}
\end{table*}

%% file: tables/best_comp_fama_tau_bench.tex
\begin{table}[h]
\small
\centering
\begin{tabular}{l|ll}
\hline
\multirow{2}{*}{\textbf{Model}} & \multicolumn{2}{c}{\textbf{Best Agent Configuration}} \\
\cline{2-3}
                       & \multicolumn{1}{c}{\textbf{Airline}} & \multicolumn{1}{c}{\textbf{Retail}} \\
\hline
Qwen3-4B-Instruct
& \makecell[l]{DCE and\\Memory (K=6)}
& \makecell[l]{DCE and\\Memory (K=6)} \\
\hline
Qwen3-14B
& \makecell[l]{DCE and\\Memory (K=2)}
& \makecell[l]{Memory (K=6)} \\
\hline
Qwen3-32B
& \makecell[l]{DCE and\\Memory (K=2)}
& \makecell[l]{DCE and\\Memory (K=6)} \\
\hline
Qwen2.5-72B-Instruct
& \makecell[l]{DCE and\\Memory (K=4)}
& \makecell[l]{DCE and\\Memory (K=2)} \\
\hline
\end{tabular}
\caption{The different agents used in the FAMA framework for $\tau$-bench.}
\label{tab:best_comp_tau_bench}
\end{table}

%% file: tables/best_comp_fama_tau_trait.tex
\begin{table}[h]
\small
\centering
\begin{tabular}{l|l|l}
\hline
\multirow{2}{*}{\textbf{Model}} & \multicolumn{2}{c}{\textbf{Best Agent Configuration}} \\
\cline{2-3}
                       & \multicolumn{1}{c}{\textbf{Telehealth}} & \multicolumn{1}{c}{\textbf{Telecom}} \\
\hline
Qwen3-4B-Instruct
& \makecell[l]{DCE and\\Memory (K=2)}
& \makecell[l]{Memory (K=6)} \\
\hline
Qwen3-14B
& \makecell[l]{Memory (K=2)}
& \makecell[l]{DCE and\\Memory (K=2)} \\
\hline
Qwen3-32B
& \makecell[l]{Memory (K=6)}
& \makecell[l]{Memory (K=6)} \\
\hline
Qwen2.5-72B-Instruct
& \makecell[l]{Memory (K=4)}
& \makecell[l]{DCE and\\Memory (K=6)} \\
\hline
\end{tabular}
\caption{The different agents used in the FAMA framework for $\tau$-trait.}
\label{tab:best_comp_tau_trait}
\end{table}

%% file: tables/best_comp_fama_ace_bench.tex
\begin{table}[h]
\small
\centering
\begin{tabular}{l|l}
\hline
\textbf{Model} & \textbf{Best Agent Configuration} \\
\hline
Qwen3-4B-Instruct
& \makecell[l]{Planner and Verifier} \\
\hline
Qwen3-14B
& \makecell[l]{Planner and Verifier} \\
\hline
Qwen3-32B
& \makecell[l]{Planner and Verifier} \\
\hline
Qwen2.5-72B-Instruct
& \makecell[l]{Planner} \\
\hline
\end{tabular}
\caption{The different agents used in the FAMA framework for ACE-Bench.}
\label{tab:best_comp_ace_bench}
\end{table}

%% file: tables/thinking_model_error_frequency.tex
\begin{table}[t]
\centering
\small
\begin{tabular}{lcc}
\toprule
\textbf{Method} & \textbf{$\tau$-airline} & \textbf{$\tau$-retail} \\
\midrule
Q3-14B-Think-ReAct & 0 & 0 \\
Q3-14B-Think-FAMA  & 7 & 3 \\
Q3-14B-Think-IRMA  & 8 & 6 \\
\midrule
Q3-32B-Think-ReAct & 0 & 0 \\
Q3-32B-Think-FAMA  & 12 & 20 \\
Q3-32B-Think-IRMA  & 13 & 23 \\
\bottomrule
\end{tabular}
\caption{Frequency of context window overflows for thinking model variants. Values indicate the number of task failures (measured for all tasks across 5 runs) where the maximum token budget was exceeded. Q3 denotes the Qwen3 model family.}
\label{tab:thinking_model_error_freq}
\end{table}

%% file: tables/SR_qwen72B.tex
\begin{table}[t]
\centering
\footnotesize
\setlength{\tabcolsep}{3pt}
\begin{tabular}{lccccc}
\toprule
\multicolumn{6}{c}{\textbf{$\tau$-bench-retail}} \\
\midrule
\textbf{Model} & \textbf{Pass\textasciicircum1} & \textbf{Pass\textasciicircum2} & \textbf{Pass\textasciicircum3} & \textbf{Pass\textasciicircum4} & \textbf{Pass\textasciicircum5} \\
\midrule
ReAct & 43.47 & 32.26 & 26.69 & 23.13 & 20.86 \\
FAMA  & 44.173 & 34.26 & 30.26 & 28.17 & 26.95 \\
SR    & 42.9 & 30.7  & 25    & 21.2  & 18.2  \\
\midrule
\multicolumn{6}{c}{\textbf{$\tau$-bench-airline}} \\
\midrule
\textbf{Model} & \textbf{Pass\textasciicircum1} & \textbf{Pass\textasciicircum2} & \textbf{Pass\textasciicircum3} & \textbf{Pass\textasciicircum4} & \textbf{Pass\textasciicircum5} \\
\midrule
ReAct & 24.4 & 18.79 & 15.6 & 12.8 & 10 \\
FAMA  & 29.2 & 21.2  & 18.8 & 18   & 18 \\
SR    & 28   & 20.4  & 17.4 & 15.6 & 14 \\
\bottomrule
\end{tabular}
\caption{Pass\textasciicircum k results for Qwen2.5-72B-Instruct on $\tau$-bench datasets comparing Self-Reflection performance against FAMA and ReAct.}
\label{tab:SR_comparisons_1}
\end{table}

%% file: tables/SR_qwen32B.tex
\begin{table}[t]
\centering
\small
\setlength{\tabcolsep}{3pt}
\begin{tabular}{lccccc}
\toprule
\multicolumn{6}{c}{\textbf{$\tau$-bench-retail}} \\
\midrule
\textbf{Model} & \textbf{Pass\textasciicircum1} & \textbf{Pass\textasciicircum2} & \textbf{Pass\textasciicircum3} & \textbf{Pass\textasciicircum4} & \textbf{Pass\textasciicircum5} \\
\midrule
ReAct & 38   & 25   & 18.8 & 15.3 & 10   \\
FAMA  & 40.5 & 26.9 & 19.9 & 15.3 & 12.2 \\
SR    & 26.7 & 15.6 & 11.3 & 9.2  & 7.8  \\
\midrule
\multicolumn{6}{c}{\textbf{$\tau$-bench-airline}} \\
\midrule
\textbf{Model} & \textbf{Pass\textasciicircum1} & \textbf{Pass\textasciicircum2} & \textbf{Pass\textasciicircum3} & \textbf{Pass\textasciicircum4} & \textbf{Pass\textasciicircum5} \\
\midrule
ReAct & 30.4 & 20   & 16.2 & 14.8 & 14 \\
FAMA  & 26.8 & 20   & 18.4 & 18   & 18 \\
SR    & 25.2 & 17   & 13.2 & 10.4 & 8  \\
\bottomrule
\end{tabular}
\caption{Pass\textasciicircum k results for Qwen3-32B on $\tau$-bench.}
\end{table}

%% file: tables/SR_qwen14B.tex
\begin{table}[t]
\centering
\small
\setlength{\tabcolsep}{3pt}
\begin{tabular}{lccccc}
\toprule
\multicolumn{6}{c}{\textbf{$\tau$-bench-retail}} \\
\midrule
\textbf{Model} & \textbf{Pass\textasciicircum1} & \textbf{Pass\textasciicircum2} & \textbf{Pass\textasciicircum3} & \textbf{Pass\textasciicircum4} & \textbf{Pass\textasciicircum5} \\
\midrule
ReAct & 25.2 & 17.8 & 14.7  & 13.2  & 12.1 \\
FAMA  & 37.9 & 25.7 & 19.7  & 16.3  & 14.7 \\
SR    & 32.1 & 19.3 & 14.69 & 12.52 & 11.3 \\
\midrule
\multicolumn{6}{c}{\textbf{$\tau$-bench-airline}} \\
\midrule
\textbf{Model} & \textbf{Pass\textasciicircum1} & \textbf{Pass\textasciicircum2} & \textbf{Pass\textasciicircum3} & \textbf{Pass\textasciicircum4} & \textbf{Pass\textasciicircum5} \\
\midrule
ReAct & 18.4 & 12   & 9.4 & 8.4 & 8  \\
FAMA  & 36.8 & 23.79& 18.2& 16.6& 16 \\
SR    & 21.2 & 15.2 & 13.3& 12.4& 12 \\
\bottomrule
\end{tabular}
\caption{Pass\textasciicircum k results for Qwen3-14B on $\tau$-bench.}
\end{table}

%% file: tables/SR_qwen4B.tex
\begin{table}[t]
\centering
\small
\setlength{\tabcolsep}{3pt}
\begin{tabular}{lccccc}
\toprule
\multicolumn{6}{c}{\textbf{$\tau$-bench-retail}} \\
\midrule
\textbf{Model} & \textbf{Pass\textasciicircum1} & \textbf{Pass\textasciicircum2} & \textbf{Pass\textasciicircum3} & \textbf{Pass@4} & \textbf{Pass\textasciicircum5} \\
\midrule
ReAct & 17.22 & 12.35 & 10.61 & 9.57  & 8.7  \\
FAMA  & 34.6  & 24.1  & 19.3  & 16.3  & 13.9 \\
SR    & 31.3  & 19.3  & 14.34 & 11.82 & 10.43 \\
\midrule
\multicolumn{6}{c}{\textbf{$\tau$-bench-airline}} \\
\midrule
\textbf{Model} & \textbf{Pass\textasciicircum1} & \textbf{Pass\textasciicircum2} & \textbf{Pass\textasciicircum3} & \textbf{Pass\textasciicircum4} & \textbf{Pass\textasciicircum5} \\
\midrule
ReAct & 32   & 28   & 26.8 & 26.4  & 26 \\
FAMA  & 37.6 & 32   & 28.3 & 26.7  & 26 \\
SR    & 33.2 & 23.9 & 19.6 & 16.39 & 14 \\
\bottomrule
\end{tabular}
\caption{Pass\textasciicircum k results for Qwen3-4B-Instruct-2507 on $\tau$-bench.}
\label{tab:SR_comparisons_4}
\end{table}